\documentclass[10pt,journal,compsoc]{IEEEtran}
\ifCLASSOPTIONcompsoc
  \usepackage[nocompress]{cite}
\else
  \usepackage{cite}
\fi
\ifCLASSINFOpdf
\else
\fi
\usepackage{amsmath}
\usepackage{algorithmic}

\usepackage{array}

\usepackage{graphicx}
\usepackage{amssymb}
\usepackage{url}
\usepackage{booktabs}
\usepackage{multicol}
\usepackage{multirow}
\usepackage{subcaption}
\usepackage{color}
\usepackage{url}
\usepackage{lipsum}

\begin{document}

\title{GPA-Net:No-Reference Point Cloud Quality Assessment with Multi-task Graph Convolutional Network}
%
%
%
%

\author{Ziyu~Shan~\IEEEauthorrefmark{1}, Qi~Yang~\IEEEauthorrefmark{1}, Rui~Ye, Yujie~Zhang, Yiling~Xu~\IEEEauthorrefmark{2}, ~\IEEEmembership{Member,~IEEE}, Xiaozhong~Xu, ~\IEEEmembership{Member,~IEEE} and~Shan~Liu, ~\IEEEmembership{Fellow,~IEEE}
\IEEEcompsocitemizethanks{\IEEEcompsocthanksitem Ziyu Shan, Rui Ye, Yujie Zhang and Yiling Xu are with Cooperative Media Innovation Center, Shanghai Jiao Tong University.\protect\\
E-mail: \{shanziyu, yr991129, yujie19981026, yl.xu\} @ sjtu.edu.cn.
\IEEEcompsocthanksitem Qi Yang, Xiaozhong Xu and Shan Liu are with Tencent Media Lab.\protect\\
 E-mail: \{chinoyang, xiaozhongxu, shanl\} @ tencent.com.}
\thanks{\IEEEauthorrefmark{1} These authors contributed equally to this work.}
\thanks{\IEEEauthorrefmark{2} Corresponding author.}
}

%
%

\markboth{Journal of \LaTeX\ Class Files,~Vol.~14, No.~8, August~2015}%
{Shell \MakeLowercase{\textit{et al.}}: Bare Demo of IEEEtran.cls for Computer Society Journals}
%



\IEEEtitleabstractindextext{%
\begin{abstract}
With the rapid development of 3D vision, point cloud has become an increasingly popular 3D visual media content. Due to the irregular structure, point cloud has posed novel challenges to the related research, such as compression, transmission, rendering and quality assessment. In these latest researches, point cloud quality assessment (PCQA) has attracted wide attention due to its significant role in guiding practical applications, especially in many cases where the reference point cloud is unavailable. However, current no-reference metrics which based on prevalent deep neural network have apparent disadvantages. For example, to adapt to the irregular structure of point cloud, they require preprocessing such as voxelization and projection that introduce extra distortions, and the applied grid-kernel networks, such as Convolutional Neural Networks, fail to extract effective distortion-related features. Besides, they rarely consider the various distortion patterns and the philosophy that PCQA should exhibit shift, scaling, and rotation invariance. In this paper, we propose a novel no-reference PCQA metric named the Graph convolutional PCQA network (GPA-Net). To extract effective features for PCQA, we propose a new graph convolution kernel, i.e., GPAConv, which attentively captures the perturbation of structure and texture. Then, we propose the multi-task framework consisting of one main task (quality regression) and two auxiliary tasks (distortion type and degree predictions). Finally, we propose a coordinate normalization module to stabilize the results of GPAConv under shift, scale and rotation transformations. Experimental results on two independent databases show that GPA-Net achieves the best performance compared to the state-of-the-art no-reference PCQA metrics, even better than some full-reference metrics in some cases. The code is available at: \url{https://github.com/Slowhander/GPA-Net.git}
\end{abstract}

}

\maketitle

\IEEEdisplaynontitleabstractindextext

%
\IEEEpeerreviewmaketitle

\IEEEraisesectionheading{\section{Introduction}\label{sec:introduction}}

%
%
%
%
\IEEEPARstart{W}{ith} the rapid development of 3D vision, point clouds have been widely used in multifarious applications, such as augmented reality (AR) \cite{alexiou2017towards,schwarz2018emerging} and automatic driving \cite{zhao2020quality}. A point cloud comprises a set of scattered points in 3D space, and each point consists of a 3D geometry coordinate and attribute information, including color, reflectance or normal vector. 
In the practical applications of point clouds, the acquired raw point cloud data is usually noisy due to the limitation of 3D sensors \cite{bletterer2020local}. Besides, the post-processing stages intended for better utilization and presentation, e.g., compression and reconstruction \cite{liu2020model}, inevitably introduce distortions for point cloud, leading to the degradation of perceptual quality  \cite{su2019perceptual}. Therefore, point cloud quality assessment (PCQA) receives commensurate attention in order to provide a better quality of experience (QoE) for human perception. 


Depending on the availability of the reference point cloud, PCQA can be classified into three categories, i.e., full-reference (FR), reduced-reference (RR) and no-reference (NR). In the past, researchers in the PCQA mostly focused on the full-reference metrics, such as Point-to-point \cite{mekuria2016evaluation}, Point-to-plane (P2lane) \cite{tian2017geometric} and Plane-to-plane \cite{alexiou2018point}, PCQM \cite{meynet2020pcqm}, GraphSIM \cite{yang2020inferring}, MS-GraphSIM \cite{zhang2021ms}, PointSSIM \cite{}, MPED \cite{yang2022mped}, etc. These metrics require full access to the reference point clouds. However, in many practical situations, the reference point clouds can not be obtained due to many limitations, such as device storage and transmission bandwidth. Therefore, inferring the point cloud quality without reference is essential to these practical applications. Unfortunately, limited researches have explored the design of no-reference PCQA metrics and failed to provide satisfying performances.

\begin{figure}[!t]
    \centering
    \includegraphics[width=0.48\textwidth]{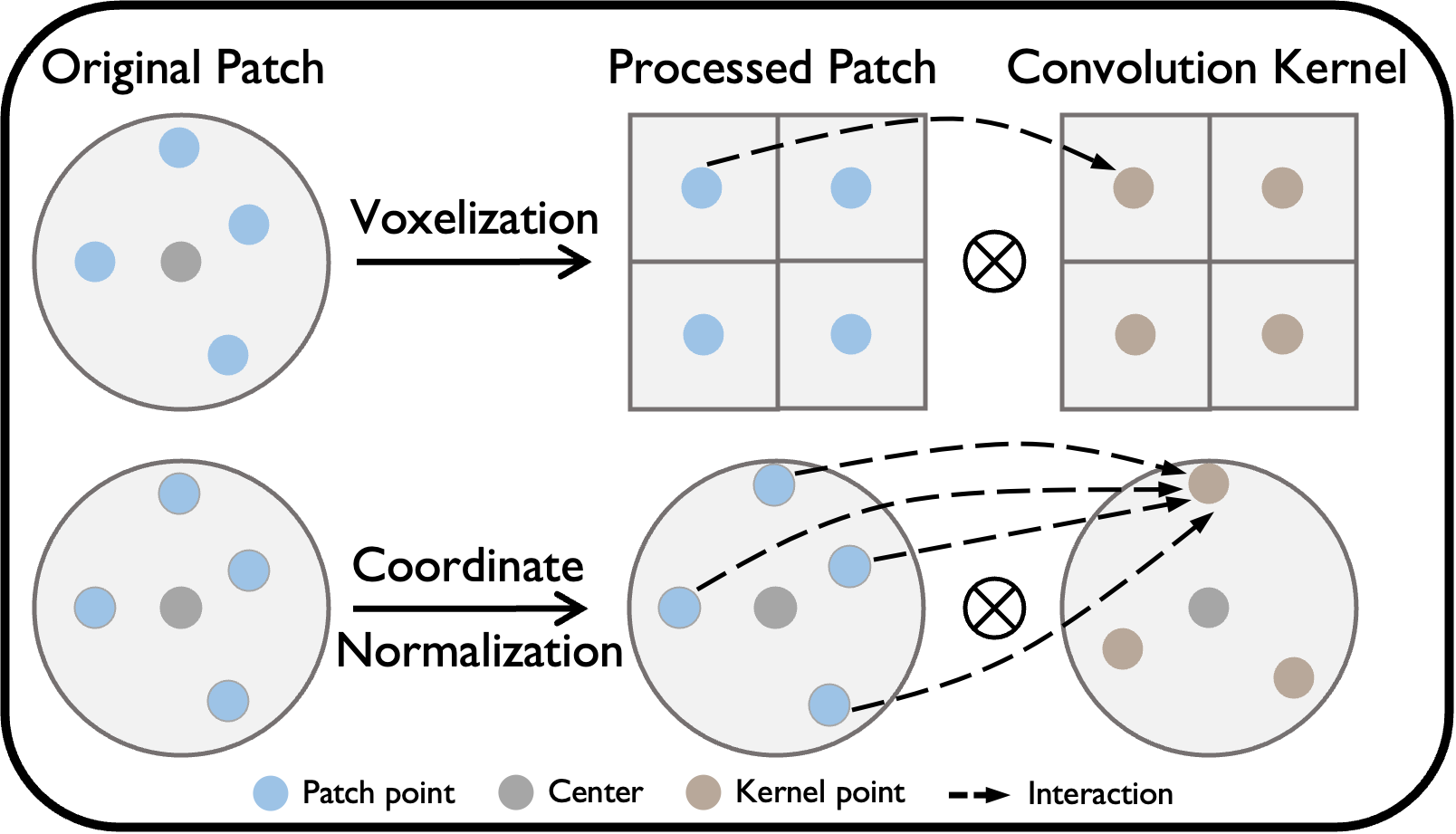}
    \caption{Illustration of preprocessing and convolution of voxelization-based metric (e.g., ResSCNN \cite{liu2022point}) and GPA-Net.}
    \label{fig:motivation}
\end{figure}

The intractability of current no-reference PCQA metrics lies in three aspects. Firstly, current metrics fail to extract effective features related to distortion properties, which means that the learned features can not fully represent the decline in point cloud quality, for two main reasons: i) extra distortion is introduced by preprocessing steps of objective metric. In these metrics, scattered point clouds are transformed to other regular formats, which introduces redundant computations and injects extra distortion to mask the effect of the original distortion. For example, the ResSCNN \cite{liu2022point} proposes using voxelization to fuse the nearby points into one voxel, which may lead to information loss and position changes. Therefore, ResSCNN fails to extract the original structural features due to the handcraft. Similar drawbacks also exist in PQA-Net \cite{RN52} and IT-PCQA \cite{yang2022no} because of multi-view projections; ii) current feature extraction networks lack the consideration of point interactions. Convolutional neural networks (CNN) are utilized for feature extraction in these no-reference metrics. However, in CNN, because of element-wise multiplication between the patch and convolution kernel, the gradient of the weight of one kernel point in backpropagation is only relevant to the spatially corresponding isolated point in the patch, as illustrated in Figure \ref{fig:motivation}. As a result, this limited point interaction cannot extract structural and textural features comprehensively \cite{liu2019relation}.

Secondly, current metrics hardly consider the impacts of various distortion types under different degrees. Generally, different distortions can bring different subjective perception results. For example, the distortion without point displacements (e.g., color noise) on the specific point cloud is more tolerable than geometric-dominated distortion (e.g., downsampling) with a relatively slight degree, and the perceptual quality declines rapidly when the color noise degree reaches a threshold \cite{watson1997digital, liu2022point}. Ignoring the above issues may limit the performance of the metrics. 

Finally, current no-reference metrics neglect the point clouds' shift, scale and rotation invariance, which leads to different objective scores for the same sample with different positions or scales. This ignorance is contrary to the fact that the observers can freely choose the point cloud position and view angle in most PCQA subjective experiments \cite{yang2020predicting, liu2021reduced, liu2022point, perry2020quality}. Ignoring these point cloud invariance may reduce the robustness of the metric.

To solve the above problems, we propose a novel and effective no-reference PCQA metric using a multi-task graph convolutional network,  namely GPA-Net. 
First, to make the network more sensitive to distortion properties, we propose a new graph convolution kernel, named GPAConv, to encode latent features on raw point cloud patches. GPAConv treats the point cloud as a graph without processing operations to preserve the original structure and texture information. Different from the state-of-the-art graph convolution kernel (e.g., 3D-GCN \cite{lin2020convolution}) that only records the most prominent feature based on point locations, the GPAConv aggregates the weighted features for point cloud patches with a deformable 3D convolution kernel, which can attentively capture the perturbation of structure and texture based on sufficient local point interactions, as illustrated in Figure \ref{fig:motivation}.
Then, to fully consider the distortion of different types and degrees, the GPA-Net treats quality score regression as the main task and predicts distortion type and degree simultaneously as the auxiliary tasks, and forces the main network to learn features that distinguish distortion of different patterns.
At last, to achieve the shift, scale and rotation invariance, we propose a coordinate normalization module. Concretely, we translate the local patches to a fixed position and normalize the scale to realize the shift and scale invariance, and then we propose simplified local reference frames (SLRF) to realize rotation invariance, which utilizes eigenvalue decomposition and simple global sign disambiguation to normalize the local orientations of point cloud patches. Therefore, compared to the previous no-reference PCQA metrics, the GPA-Net can extract the vital distortion-related features on the native point clouds more effectively and exhibit invariance to shift, scaling and rotation, demonstrating the superiority in terms of both effectiveness and robustness.

The main contributions of our work are as follows:
\begin{itemize}
    \item We propose a multi-task graph convolutional network, GPA-Net, to realize no-reference PCQA. We infer point cloud quality by considering point cloud content (structure and texture), distortion type, and degree jointly, which can better quantify the influence of distortion. 
    
    \item We propose a novel graph convolution method, GPAConv, which attentively captures the perturbation of structure and texture based on local point interactions. 
    
    \item We propose a local coordinate normalization module featuring an efficient local reference frame (SLRF) to make our model invariant to shift, scaling and rotation.
    
    \item We compare the proposed GPA-Net with multiple full-reference and state-of-the-art no-reference PCQA metrics. Experimental results show that our GPA-Net outperforms all the no-reference and majority full-reference metrics. Ablation studies reveal that GPA-Net presents impressive robustness to the perturbations, i.e., translation, scaling, and rotation.
\end{itemize}

The remainder of this paper is structured as follows. In section \ref{sec:related}, we review the related works on no-reference PCQA metrics, convolution on point clouds and multi-task learning. In section \ref{sec:gpaconv}, we introduce the proposed GPAConv in detail. In section \ref{sec:gpanet}, we reveal the proposed GPA-Net, including the SLRF. In section \ref{experiment}, we illustrate the experiment results of the proposed metric and other state-of-the-art metrics.  Finally, in section \ref{sec:conclusion}, we conclude this paper.

\section{Related Work}\label{sec:related}
In this section, we briefly review the literature related to our approach, including no-reference PCQA metrics, convolution on point clouds and multi-task learning.

\textbf{No-reference point cloud quality assessment.} Due to prior knowledge and database limitations, limited no-reference quality assessment metrics for point cloud have been proposed.
Zhang et al. \cite{zhang2022no} propose a no-reference PCQA metric for colored based on some handcrafted features. Point clouds are first projected from 3D space into geometry and color feature domains, then multiple statistics features are extracted to achieve quality prediction base on one support vector machine (SVM). However, this traditional metric performs much worse compared to the learning-based models.

Liu et al. \cite{liu2022point} propose a learning-based no-reference PCQA network, named ResSCNN, using voxelization followed by sparse 3D-CNN. 
Chetouani et al. \cite{chetouani2021deep} propose a model using traditional CNN networks (e.g., VGG \cite{simonyan2014very}) that first extract three low-level features (i.e., geometric distance, local curvature, and luminance values) from local patches and feed them into CNN.
Liu et al. \cite{RN52} propose PQA-Net using multi-view projection followed by 2D CNN and  distortion type identification module, but it is not end-to-end trained and ineffective when superposed distortion exists.
Fan et al. \cite{fan2022no} propose a model that transforms point cloud into three captured video sequences by rotating the camera around the observed point cloud through three orbits, then uses ResNet3D to extract spatial and temporal information jointly. 
These learning-based models transform irregular point clouds into regular patterns such as voxels and extract features with CNN. However, these approaches introduce unnecessary computations and undesired information loss. Besides, because of the non-deformable convolution kernel, CNN is not suitable for the irregular structure of point clouds, which fails to capture critical distortion-related features.

Tliba et al. \cite{tliba2022point} employ PointNet \cite{qi2017pointnet} as backbone to extract patch-wise local features and train the network in both supervised and unsupervised ways. However, due to the symmetric pooling operation, this point-based network can not explicitly encode the local geometric relationships between neighbors, and point clouds' shift, scaling and rotation invariance are not considered.

\textbf{Convolution on point clouds.} Many methods have been proposed to define a proper convolution on point clouds. The state-of-the-art methods that retain the structure of point clouds can be divided into two categories: point-based and graph-based methods. 

Most of the point-based model the point cloud convolution as set-input problems and use symmetric pooling to realize permutation and cardinality invariance. PointNet \cite{qi2017pointnet} is the first point-based method, which uses weight-shared MLPs on each point to aggregate global features. PointNet++ \cite{qi2017pointnet++} proposes to apply PointNet layers locally in a hierarchical architecture. KPConv \cite{thomas2019kpconv} proposes a deformable convolution using explicit and shift-learnable kernel points. Li et al. \cite{li2021rotation} propose a rotation-invariant network using a new local representation and introduce a region relation convolution to encode non-local information to alleviate inevitable global information loss caused by the normalized orientations. However, in these point-based methods, point interaction is insufficient, resulting in a bad performance in structural feature extraction. 

Graph is a natural representation for point clouds to model local structures or textures, where the edge features in a graph can naturally describe the relationships between neighbors in terms of geometry and attribute. DGCNN \cite{wang2019dynamic} proposes a simple operation, called EdgeConv, which captures local geometric structure while maintaining permutation invariance by generating edge features that describe the relationships between a point and its neighbors. GAC \cite{wang2019graph} proposes a sharing attention mechanism to adapt to the structure of objects dynamically. 3D-GCN \cite{lin2020convolution} proposes a deformable kernel with shift and scale-invariant properties. AdaptiveConv \cite{zhou2021adaptive} proposes an adaptive convolution kernel to produce features that are more flexible to shape geometric structures. 

However, most of point-based and graph-based methods are all variant to the rotation of the input point cloud, and they may not perform well on dense point clouds since they are all designed for classification or segmentation of simple structured and textured point clouds \cite{chang2015shapenet,wu20153d}.

\textbf{Multi-task learning.} Multi-task learning (MTL) aims to learn multiple related tasks jointly so that the knowledge contained in a task can be leveraged by other tasks \cite{zhang2021survey}. MTL has been widely applied in image quality assessment (IQA) because related auxiliary tasks can force the network to learn more comprehensive features that are sensitive to distortion characteristics.
Xu et al. \cite{xu2016multi} proposes an MTL framework to train multiple IQA models together, where each model is for each distortion type. The proposed model can exploit the common features among various distortion patterns and prevent overfitting problem in distortion-aware metrics.
Golestaneh et al. \cite{golestaneh2020no} uses distortion types as well as subjective human scores to predict the image quality, which are fused by a proposed feature fusion network. 
Hou et al. \cite{hou2020content} proposes an MTL network with two weight-shared CNN branches to find the quality differences and reduce the content dependency caused by the insufficient training data, which is a common case in quality assessment. 
Li et al. \cite{li2022blind} deconstructs IQA into three elements. i.e., image content, degradation pattern and intensity and designs the corresponding auxiliary tasks. In addition, \cite{li2022blind} exploits the progressive relevance among these tasks and introduce it into the multi-task learning framework to constrain the hypothesis space of the main task.
However, to the best of our knowledge, none of the existing works investigate the effect of MTL on PCQA.
\section{GPAConv: Structure and Texture Attentive Graph Convolution}\label{sec:gpaconv}
In this section, we introduce the proposed 3D deformable convolution GPAConv in detail, which can perform convolution in irregular spatial points and extract effective structural and textural patterns. To better illustrate the convolution process, we take an input point cloud patch as an example, and demonstrate the superiority of GPAConv by employing a toy example. 

We assume that the input cloud patch consists of a center point $x_0$ and neighbors, which is consistent with our GPA-Net in Section \ref{sec:gpanet}. Specifically, we denote the input point cloud patch whose center point is $x_0$ as $\mathcal{N}_{x_0}^K = \{ x_k |\forall x_k \in KNN(x_0, K),k=0,1,...,K\}$, where a point $x_k = [p_k, f_k]$, $p_k \in \mathbb R^3$ represents the coordinate and $f_k \in \mathbb R^D$ represents the $D$-dimensional feature, $KNN(x_0, K)$ denotes the $K$ nearest neighbors of point $x_0$. 


Mathematically, referring to 3D-GCN \cite{lin2020convolution},  we define the convolution kernel as $\mathcal S = \{s_j|j=0,1,...,J \}$, where $s_0$ is the kernel center and $s_1$ to $s_J$ represent $J$ supports. Note that $J$ is much less than $K$ \cite{lin2020convolution}.  The coordinate of $s_0$ is set as $d_0 = (0, 0, 0)$.  As for the $J$ supports,  we define them as $s_j = [d_j, w_j]$, where $ d_j \in \mathbb R^3$ represents the coordinate of the support corresponding to $s_0$ and $w_j \in \mathbb R^D$ is the learned weighting parameters. For supports $s_1$ to $s_{J}$, their coordinates also represent their direction vectors because the center is set at the origin.

For 2D CNN, the convolution operation can be regarded as computing the similarity between the patch and the 2D convolution kernel, i.e., the inner product. Generally, the greater the inner product comes with the higher similarity. Similarly, we define the convolution operation for 3D point clouds as computing the similarity between the patch $\mathcal N_{x_0}^K$ and the kernel $\mathcal S$. However, for point clouds whose points are scattered in the space, there is no definite correspondence between patch points and kernel supports. To address this problem, we compute the inner product between each $f_k$ and $w_j$ with $k,j \neq 0$ to quantify feature similarity, and use geometric similarity to weight the feature similarity to produce the final results. Mathematically, for the patch $\mathcal N_{x_0}^K$ and the kernel $\mathcal S$, the convolution operation $\text{GPAConv}(\mathcal N_{x_0}^K, \mathcal S)$ is defined as:
\begin{equation}\label{eq:gpaconv}
    {\text{GPAConv}(\mathcal N_{x_0}^K, \mathcal S)} = \sigma\Big(\langle f_0, w_0 \rangle + \sum_{j=1}^{J} \sum_{k=1}^{K} g_{j,k} \cdot \langle f_k, w_j \rangle\Big)
\end{equation}
where $\langle \cdot, \cdot\rangle$ denotes the inner product of two vectors; $\sigma$ is the nonlinear activation function; $g_{j,k}$ is the weight factor determined by the geometric similarity of $x_k$ and $s_j$, which is defined as:
\begin{equation}
    g_{j,k} = \mathop{\text{softmax}}\limits_k \bigg(\gamma \cdot  \frac{ \langle p_k, d_j \rangle}{||p_k|| \cdot ||d_j||} \cdot \min \Big( \frac{||p_k||}{||d_j||} , \frac{||d_j||}{||p_k||} \Big) \bigg),
\label{eq:g}
\end{equation}
in which $||\cdot||$ denotes the length of the vector and $\gamma$ is a learnable scaling parameter. 
Here we split the geometric similarity into length similarity and direction similarity. 
The linear ratio between the lengths of the point direction vector and support vector is used as length similarity and their cosine value is used as direction similarity. The direction similarity is in [-1,1] and equals 1 or -1 when $p_k$ and $d_j$ are in the same or opposite direction, respectively. 
The min($\cdot$) operation can realize the symmetric property and scale the length similarity to (0,1], while $\gamma$ before $\text{softmax}$ operation is used to control the assigned weight of each patch point and fix possible vanishing gradient problem introduced by softmax function \cite{vaswani2017attention}. {\textit{Permutation invariance}} is realized by weighted summation of the point-level pair-wise similarity between all the patch points and supports.Specifically, if the data feeding order changes, the order of $g_{j,k}$ changes accordingly, and the result remains the same.
\begin{figure}
    \centering
    \includegraphics[width=0.48\textwidth]{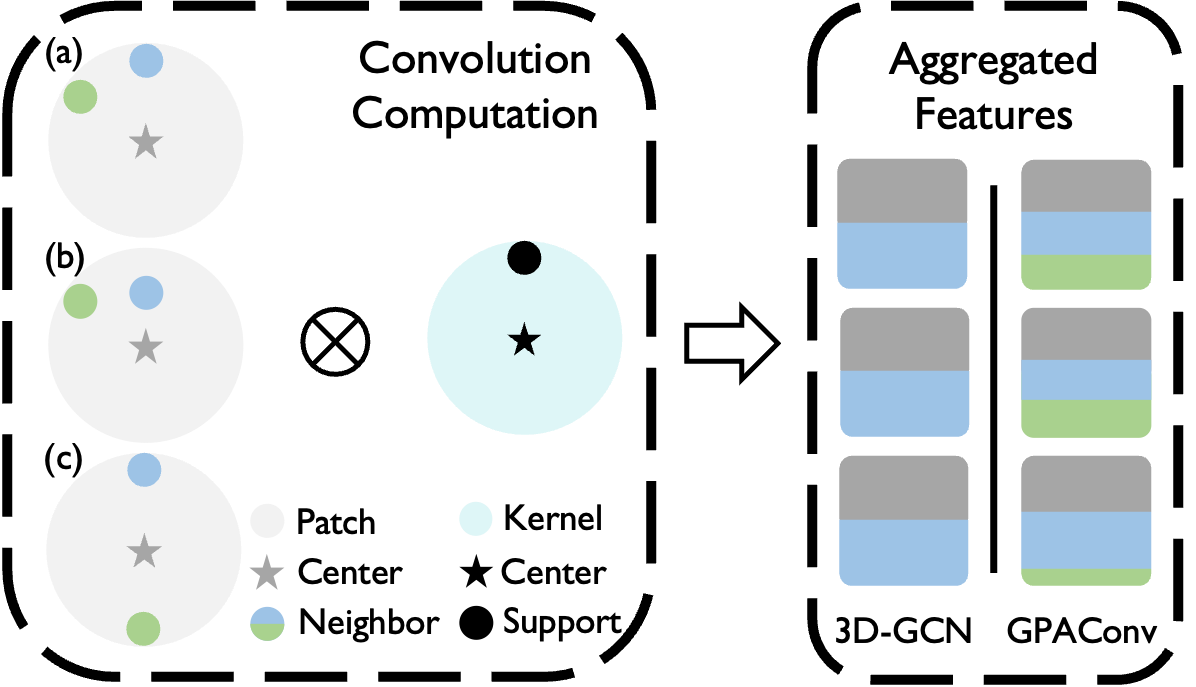}
    \caption{A toy example of convolution computations for 3D-GCN and the proposed GPAConv.}
    \label{fig:toy}
\end{figure}
\begin{figure*}
    \centering
    \includegraphics[width = \textwidth]{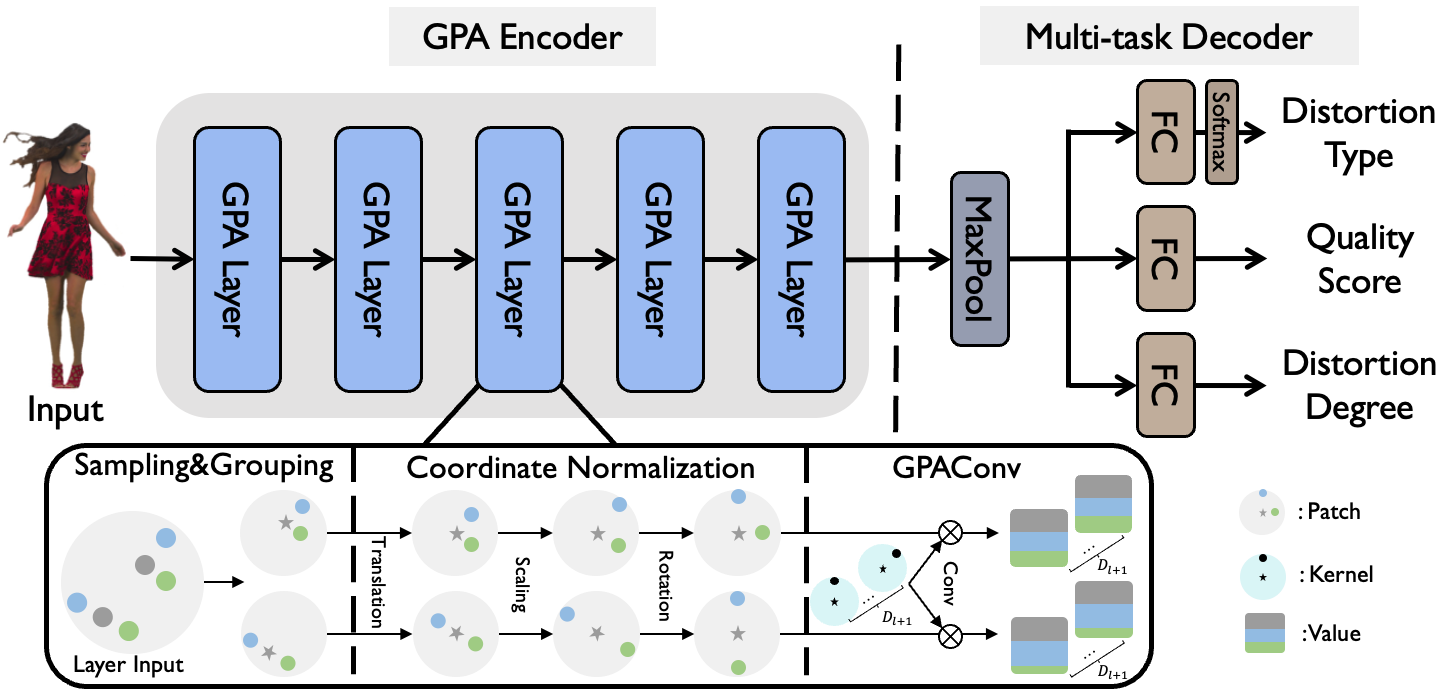}
    \caption{Illustration of our proposed GPA-Net, which consists of 2 parts, i.e., the GPA encoder and multi-task decoder. (a) The GPA encoder consists of 5 GPA layers, and each GPA layer contains 3 modules, i.e., the sampling\&grouping module, the coordinate normalization module and the GPAConv module. (b) The multi-task decoder consists of a channel-wise max-pooling module and 3 fully-connected layers.}
    \label{fig:arch}
\end{figure*}

{\bf Analysis:} We demonstrate the effectiveness of GPAConv by comparing it with the kernel used in 3D-GCN. 3D-GCN computes the similarity by $\langle f_0, w_0 \rangle + \sum_{j=1}^{J} {\displaystyle{ \max_k}} \langle f_k, w_j \rangle \cdot \frac{\langle p_k, d_j \rangle}{||p_k|| \cdot ||d_j||}$, which only utilizes the direction similarity. Meanwhile, the similarity is only contributed by the patch center and a single neighbor for each kernel point due to the $\max(\cdot)$ operation. The ignorance of length similarity and the $\max(\cdot)$ operation may cause geometry and attribute information loss, which is severe for practical situations where dense point clouds are processed.

Compared to 3D-GCN, our GPAConv follows Eq. \eqref{eq:gpaconv} and utilizes both direction and length similarity. Also, its similarity is attentively contributed by the patch center convolution and all neighbors convolution. Therefore, GPAConv can capture the geometrical and attribute characteristics more effectively and exhaustively.

We demonstrate a toy example in Figure \ref{fig:toy}, in which we take three patches with different patterns and a fixed convolution kernel as example. Besides a center, each patch has two neighbors, which are denoted by blue and green point whose features are $f_1$ and $f_2$. Taking patch (a) as reference, patch (b) shares the same green point, but the blue point shrinks towards to the center; patch (c) share the same blue point but different positions of green point. According to the formulation of kernel convolution in 3D-GCN, the results of 3D-GCN over the three different patches are identically $\langle f_0, w_0 \rangle + \langle f_1, w_1 \rangle$, where we assume that $\langle f_1, w_1 \rangle \geq \langle f_2, w_1 \rangle$. Here the feature of green point ($f_2$) and the shrinkage of blue point ($f_1$) are neglected, see the identical aggregated features in Figure \ref{fig:toy}.

In contrast, the result of our GPAConv is $\langle f_0, w_0 \rangle + g_{1,1} \langle f_1, w_1 \rangle + g_{1,2} \langle f_2, w_1 \rangle$, where $g_{1,1}$  and $g_{1,2}$ vary across the three patches. For example, the $g_{1,1}$ for patch (a) is greater than patch (b) because the blue point in patch (a) is more ``close'' to the support ($w_1$) in terms of lengths. Here the displacement of green point ($f_2$) and the shrinkage of blue point ($f_1$) are all perceived attentively, and the complete texture formed by $f_1$ and $f_2$ are considered, see discriminative aggregated features in Figure \ref{fig:toy}. Also, this superiority is more significant for dense clouds which consist of hundreds of points in a patch.

{\bf Conclusion:} In summary, the proposed GPAConv can attentively capture the perturbation of structure and texture, improving the sensibility to the point cloud distortion detection. Moreover, the GPAConv demonstrate the superiority in a more distinct manner in large-scale situations where the local point cloud patches can be considered as toy examples with higher point density.

\section{GPA-Net: Graph Convolutional Network with Multi-task Learning}\label{sec:gpanet}
In this section, we introduce the proposed no-reference metric, i.e., GPA-Net, in detail. The Framework of GPA-Net is shown in Figure \ref{fig:arch}, which consists of two main parts, the GPA encoder and multi-task decoder. The GPA encoder transforms point clouds into high-dimensional latent features and the multi-task decoder utilizes the features to predict quality score, distortion type and degree. 

\subsection{GPA Encoder}\label{sec:layer}
As illustrated in Figure \ref{fig:arch}, the proposed GPA encoder consists of 5 cascaded GPA layers, and each GPA layer contains three modules, i.e., sampling and grouping, coordinate normalization and GPAConv. Specifically, the sampling and grouping module splits the input point clouds into multiple patches; the coordinate normalization module normalizes the local patches to a fixed position and scale to realize shift, scale and rotation invariance; the GPAConv module utilizes the proposed GPAConv kernels in Section \ref{sec:gpaconv} to generate latent features for each patch. We will introduce these three modules carefully in the following parts.
 
\subsubsection{Sampling and grouping}
To extract local features efficiently, we use random sampling to generate keypoints, take the keypoints as centers and use geometry-based KNN search to construct local patches. Considering dense point clouds have relatively uniform point densities, KNN is preferred in our work rather than other methods (e.g., ball-query grouping \cite{qi2017pointnet++} and inverse density scale grouping \cite{wu2019pointconv}) due to its low complexity.

Mathematically, for the input point cloud in the $l$-th layer $X_l = [P_l, F_l]\in \mathbb{R}^{3+D_l}$ with $D_l$-dimension feature $F_l$ and coordinates $P_l$, the sampling and grouping (SG) can be formulated as:
\begin{align}
\begin{split}
    &\text{SG}(X_l, \theta_l) = \mathcal N_{X_{l,\phi}}^{K_l} = [\mathcal P_{l}, \mathcal F_{l}] \in \mathbb{R}^{(1+K_l)\times(3+D_l)} \\
\end{split}
\end{align}
where  $\mathcal N_{X_{l,\phi}}^{K_l}$ represents the set of generated patches with keypoints as $X_{l,\phi} = [P_{l,\phi}, F_{l,\phi}] = \{x_{l,0}^{j}|j=1,2,...,q_l\}$; $q_l$ and $K_l$ are respectively the number of keypoints and the neighborhood size in layer $l$; $\mathcal P_l$, $\mathcal F_l$ is the coordinates and features of $\mathcal N_{X_{l,\phi}}^{K_l}$, respectively; $\theta_l$ is the set of sampling and grouping parameters, including the values of $K_l$ and $q_l$, which are chosen carefully to ensure a slight overlap between each patch and good space coverage.

\subsubsection{Coordinate normalization}
After sampling and grouping, we have a series of point cloud patches. To realize shift, scale and rotation invariance, we use coordinate normalization to adjust the spatial coordinates of patch points.

Given the patch $\mathcal N_{x_0}^{K}$ with coordinates $\mathcal P$ and features $\mathcal F$ in a random layer of GPA encoder with the keypoint $x_0$ and neighborhood size $K$, we respectively consider the above three invariance. 
Firstly, for the shift invariance, we translate the patch to set the coordinate of $x_0$ as origin, i.e., $p_0 = (0, 0, 0)$, which transforms the coordinates of other points in $N_{x_0}^{K}$ into the relative coordinates corresponding to $p_0$ and eliminate the influence of possible shift.
Secondly, for the scale invariance, we use scaling factors denoted as $\alpha$ that are shared in an individual patch to normalize the distances of all the patch points corresponding to $x_0$ into the range $(0, 1]$. 

To realize the rotation invariance, we propose a simplified local reference frame (LRF) for the patches, i.e., SLRF. We define the coordinate system of original point clouds as global reference frame,  while LRF is a local coordinate system (with three orthogonal axes) built on a local patch surface, which is considered as repeatable under rigid rotations \cite{ao2020repeatable, bro2008resolving,salti2014shot}. We rotate the patches to a fixed position by the rotation matrix between SLRF and global reference frame. 

Specifically, the sketch of LRF can be divied into three steps: i) the barycenter $\hat{x}$ of the $K$ nearest neighbors of $x_0$ is first calculated; ii) the eigenvectors of covariance matrix $\mathbf{\Sigma_{\hat{p}}} = \frac{1}{K} \Sigma_{i=0}^K (p_i - \hat{p})(p_i - \hat{p})^T$ are calculated and denoted as $[v_1,v_2,v_3]$ in decreasing eigenvalue order. $[v_1,v_2,v_3]$ are regarded as $x,y$ and $z$ axis of the LRF , where $z$ axis is the estimated normal \cite{bro2008resolving}; iii) for the uniqueness of LRF, the direction that the majority of the $K$ nearest neighbors orient to the neighborhood center is used to disambiguate the sign ambiguity of the eigenvectors. However, for dense point clouds, which usually exist numerous patches and each patch may have hundreds of points, the traditional LRF is very time-consuming for two reasons: i) barycenter computations; ii) all the patch points are considered for $\Sigma_{\hat p}$  computation and local disambiguation.

In the proposed SLRF, we mainly improve the traditional LRF in terms of computation complexity. Specifically, we neglect the barycenter and replace it with the keypoint $x_0$ referring to \cite{salti2014shot}. Besides, we use the $M$ ($M<K$) nearest neighbors of $x_0$ to disambiguate the sign ambiguity. Specifically, we derive a more straightforward covariance matrix:
\begin{equation}
    \mathbf{\Sigma}_{p_0} = \frac{1}{M} \sum_{i=1}^M \alpha^2(p_i-p_0)(p_i-p_0)^T
\end{equation}

Then, we compute the directions of the axis using singular value decomposition (SVD) or eigenvalue decomposition (EVD) of the covariance matrix. Due to the sign ambiguity of eigenvectors, the estimated normal may orient to the internal of the point cloud rather than the external. Therefore, we propose a simple global sign disambiguation method to reorient the axes:
\begin{equation}
    z = \left \{ 
    \begin{aligned}
    v_3, & \quad \langle v_3, p_c-p_0 \rangle \geq 0 \\
    -v_3, & \quad \langle v_3, p_c-p_0 \rangle < 0 \\
    \end{aligned}
    \right.
\end{equation}
where we set the eigenvectors in decreasing eigenvalue order as $v_1, v_2, v_3$, respectively. $p_c$ is the coordinate of the spatial center of the point cloud bounding box. This disambiguation ensures that the directions of the axes are consistent with $p_c-p_0$. $x$ axis is disambiguated similar to $z$ axis, and $y$ axis is obtained via $z$ and $x$. In this disambiguation, inner product  is computed only once between each estimated axis and the patch center's global direction vector, which reduces computation complexity compared to the local disambiguation of LRF.

Finally, we denote the $x, y, z$ axes of patches as local axes and the $x, y, z$ axes of global reference frame as global axes. Our goal is to unify the normal (local $z$ axis) orientation of all the patches, and it is feasible to rotate the local $z$ to any global axis. Here we rotate the local $z, x$ axis to let them orient to the global $y, z$ axis, respectively, and we will compare the results of other different axis rotation strategies in section \ref{sec:ab_lrf}. After the SLRF-based reorientations,  the rotation invariance is achieved.

Now we have achieved all three invariance. The overall coordinate normalization can be formulated as:
\begin{align}
\begin{split}
    {\rm CoordNorm}(\mathcal{N}_{x_0}^{K})=& [R (S \odot (\mathcal P+T)), \mathcal F]\\
    = &\widetilde {\mathcal N}_{x_0}^{K}=[\widetilde{\mathcal P}, \mathcal F], 
    \end{split}
\end{align}
where $\widetilde {\mathcal N}_{x_0}^{K}$ is the normalized patches with the normalized coordinates $\widetilde{\mathcal P}$, $T$ is the translation matrix for the patches. The operator $\odot$ denotes element-wise product; $S$ represents scaling operator; $R$ represents rotation matrix computed as \cite{kovacs2012rotation}.

\subsubsection{GPAConv module.}
After the coordinate normalization, we utilize GPAConv kernel to convolve the normalized patches in layer $l$ to get the latent features, as well as the input features of the next layer, i.e., $F_{l+1}$. Then we concatenate the absolute coordinates of keypoints $X_{l, \phi} $ and $F_{l+1}$ to generate the new point cloud with aggregated features and feed it to the next layer.

The GPAConv operation in layer $l$ can be formulated as:
\begin{align}
\begin{split}
    &\text{GPAConv}(\widetilde{\mathcal N}_{X_{l,\phi}}^{K_l}, \mathcal S_l) =  F_{l+1}, \\
    &X_{l+1}=[P_{l,\phi}, F_{l+1}]\in \mathbb R^{q_l\times(3+D_{l+1})},
\end{split}
\end{align}
where $X_{l+1} $ is the input point cloud of layer $(l+1)$.
\begin{table*}[]
    \centering
    \caption{Comparison of GPA-Net with other PCQA metrics for SJTU-PCQA database.}
    \begin{tabular}{ccccccc|cccc}
    \toprule
        &\multicolumn{6}{c}{Full-Reference} & \multicolumn{4}{c}{No-Reference}\\
        \midrule
        & M-p2po & M-p2pl & H-p2po & H-p2pl & PCQM & GraphSIM & ResSCNN & IT-PCQA & PQA-Net & GPA-Net\\
        \midrule
        SROCC & 0.718 & 0.667 & 0.583 & 0.598 & 0.846 &  \textbf{0.908} & 0.834 & 0.597 & 0.689 & \textbf{0.875}\\
        PLCC & 0.669 & 0.627 & 0.599 & 0.613 &  0.860 & \textbf{0.893} & 0.863 & 0.618 & 0.705 & \textbf{0.886}\\
    \bottomrule
    \end{tabular}
    \label{tab:sjtu}
\end{table*}

\begin{table*}[]
    \centering
    \caption{Comparison of GPA-Net with other PCQA metrics for LS-PCQA database.}
    \begin{tabular}{ccccccc|cccc}
    \toprule
        &\multicolumn{6}{c}{Full-Reference} & \multicolumn{4}{c}{No-Reference}\\
        \midrule
        & M-p2po & M-p2pl & H-p2po & H-p2pl & PCQM & GraphSIM & ResSCNN & IT-PCQA & PQA-Net & GPA-Net\\
        \midrule 
        SROCC & 0.301 & 0.291 & 0.262 & 0.264 & \textbf{0.537} & 0.460 & 0.594 & 0.553 & 0.588 & \textbf{0.602}\\
        PLCC & \textbf{0.611} & 0.589 & 0.525 & 0.607 & 0.398 & 0.427 & 0.624 & 0.534 & 0.592 &  \textbf{0.628}\\
    \bottomrule
    \end{tabular}
    \label{tab:ls}
\end{table*}

\begin{table*}[]
    \centering
    \caption{Comparison of GPA-Net with other PCQA metrics for WPC database}
    \begin{tabular}{ccccccc|cccc}
    \toprule
        &\multicolumn{6}{c}{Full-Reference} & \multicolumn{4}{c}{No-Reference}\\
        \midrule
        & M-p2po & M-p2pl & H-p2po & H-p2pl & PCQM & GraphSIM & ResSCNN  & IT-PCQA  & PQA-Net & GPA-Net(ours)\\
        \midrule
        SROCC & 0.566 & 0.446 & 0.258 & 0.315 & \textbf{0.743} &  0.680 & 0.735 & 0.556 & 0.698 & \textbf{0.758}\\
        PLCC & 0.578 & 0.488 & 0.398 & 0.383 &  \textbf{0.751} & 0.694 & 0.752 & 0.574 & 0.712 & \textbf{0.769}\\
    \bottomrule
    \end{tabular}
    \label{tab:wpc}
\end{table*}

\subsection{Multi-task Decoder} \label{sec:arch}
As illustrated in Figure \ref{fig:arch}, the multi-task decoder consists of 2 main parts: global max-pooling and 3 paratactic fully-connected layers. 

The global max-pooling module (GMP) applies channel-wise max-pooling to generate the final feature, i.e., $F_{out}$, 
\begin{equation}
    F_{out}={\rm GMP}(F_5). 
\end{equation}

Then, the 3 fully-connected layers respectively achieve the main task, i.e., quality score regression and the auxiliary tasks, i.e.,  distortion type classification and distortion degree regression. Mathematically, we define the multi-task decoder as $\vec{D}$, the decode procedure is formulated as:
\begin{align}
\begin{split}
    &\vec{D}(F_{out}) = [\hat M, \vec {P}, \hat{D}]\\
    &\hat{P} = \text{softmax}(\vec{P})
    \end{split}
\end{align}
where $\hat M$ is the predicted quality score, $\hat D$ is the predicted distortion degree, $\hat P = \{\hat p_1, {\hat p_2}, ..., \hat p_C\}$ is the predicted probability over $C$ distortion types, and $\vec{P}$ is output of the fully-connected layer for distortion type prediction before softmax. With the two auxiliary tasks, GPA-Net can learn more distortion-related features and achieve better performance in predicting quality score. Note that distortion degree is linearly normalized to adapt to the distortion settings of different datasets.

\subsection{Loss Function} \label{sec:loss}
To condition the training of the different tasks of GPA-Net, we use 3 different loss functions for the decoder. Given the ground truth mean opinion score (MOS), distortion type and degree as $M, P, D$, respectively, where $P$ is one-hot pattern, the loss functions are defined as:
\begin{equation}
    \mathcal{L}_1= \left \{ 
    \begin{aligned}
    &0.5(E-\hat E)^2  \quad \text{ if } |E-\hat E| < 1 \\
    &|E-\hat E| - 0.5  \quad \text{otherwise} \\
    \end{aligned}
    \right.
\end{equation}

\begin{equation}
    \mathcal{L}_2 = -\sum_{i=1}^C p_i \log(\hat p_i) + (1-p_i)\log(1-\hat p_i)
\end{equation}
where smoothL1 loss is used for MOS regression and distortion degree regression, thus $\mathcal{L}_1$ and $\mathcal{L}_2$ share the same form. Cross-entropy loss $\mathcal L_2$ is used for distortion type classification. The total loss is defined as the weighted sum of these loss functions:

\begin{equation}
    \mathcal{L} = \lambda_1 \mathcal{L}_1 + \lambda_2 \mathcal{L}_2 +\lambda_3 \mathcal{L}_3
\end{equation}

To control the weights of these tasks (i.e., $\lambda_1$, $\lambda_2$ and $\lambda_3$), the dynamic weight averaging (DWA) strategy \cite{liu2019end} is adopted, which is a popular approach to balance the contribution of different tasks in multi-task learning by dynamically adjusting the weight of each task according to the loss convergence rate. The dynamic weight of the task $k$ in training epoch $t$ is defined as:
\begin{equation}
    \lambda_k = \frac{3\exp (w_k(t-1))}{\sum_{i=1}^3 \exp (w_i(t-1))}, \quad k=1,2,3
\end{equation}
where $w_k(t-1)$ denotes the loss convergence rate of task $k$ in epoch $t-1$, which is defined as:
\begin{equation}
   w_k(t-1) = \frac{\mathcal{L}_k(t-1)}{\mathcal{L}_k(t-2)}, \quad k=1,2,3
\end{equation}

By adopting the DWA strategy, the GPA-Net can automatically select the most appropriate weight based on the convergence rate of each task-specific loss, thereby replacing manual parameter tuning and improving the robustness of model training. Note that distortion degree is normalized for different datasets.

\section{Experiments}\label{experiment}
\subsection{Dataset Preparation}
In the experiments, three PCQA databases SJTU-PCQA \cite{yang2020predicting}, LS-PCQA\cite{liu2022point}, WPC \cite{liu2022perceptual} are employed to test the performance of the proposed model.
\begin{itemize}
    \item \textbf{SJTU-PCQA.} The SJTU-PCQA \cite{yang2020predicting} database contains 9 native point cloud samples and 378 distorted point cloud samples. Each native reference point cloud is impaired with 7 different types of distortions under 6 levels, including octree-based compression, color noise, geometry gaussian noise, downsampling, and 3 superimposed distortions: downsampling and color noise, downsampling and geometry gaussian noise, color noise and geometry gaussian noise.
    \item \textbf{LS-PCQA.} The LS-PCQA \cite{liu2022point} database is a large-scale PCQA database, which contains 104 high quality native point clouds and 24,024 distorted point clouds, 1,320 of them providing MOS. Each native reference point cloud is impaired with 33 types of distortions (e.g., V-PCC, G-PCC, geometry noise, contrast noise, luminance noise) under 7 levels.
    \item \textbf{WPC.} The WPC database \cite{liu2022perceptual} contains 20 native point cloud samples and 740 distorted samples. Each native reference point cloud sample is impaired with 4 different types of distortions, including downsampling, gaussian noise, G-PCC and V-PCC.
\end{itemize}

\begin{figure*}[h]
        \centering
        \begin{subfigure}[b]{0.2\textwidth}
            \centering
            \includegraphics[width=4cm]{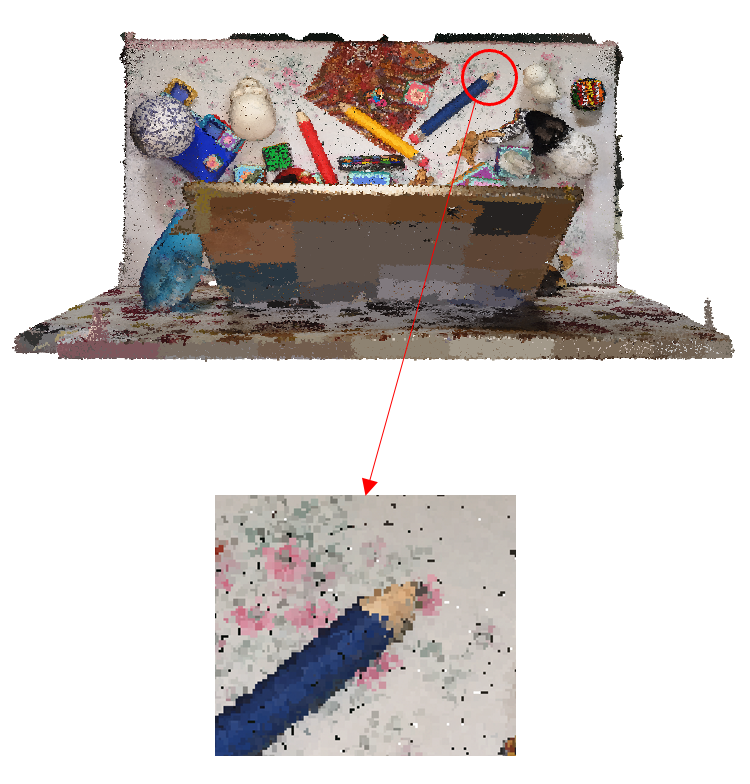}
            \vspace{0.2cm}
            \caption[]%
            {\centering{\small MOS:4.63, DGN, 3, score:(5.21 5.45 4.99  {\color{blue}{4.86}})}}    
            \label{fig:qua_1}
        \end{subfigure}
        \hspace{0.7cm}
        \begin{subfigure}[b]{0.2\textwidth}  
            \centering 
            \includegraphics[width=4cm]{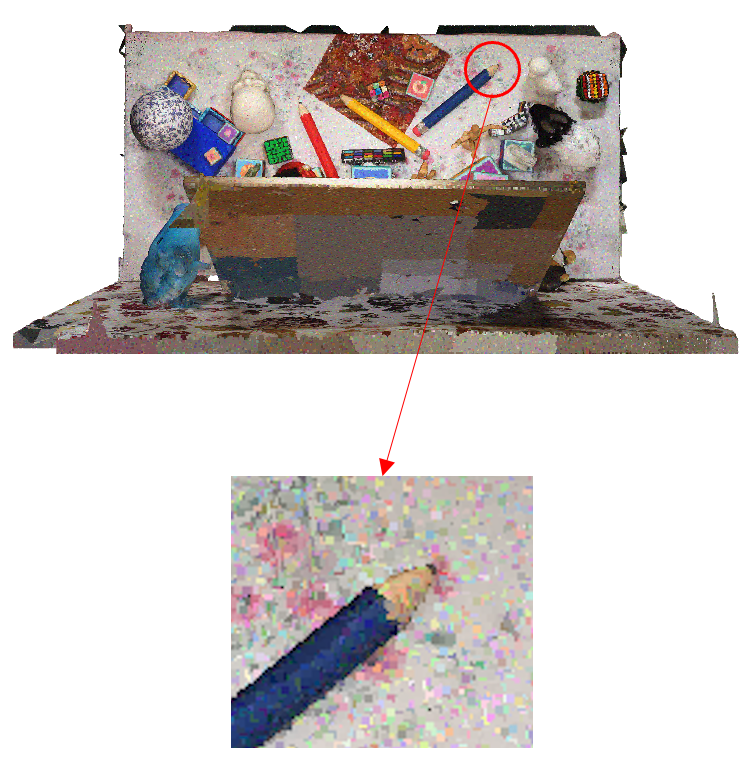}
            \vspace{0.2cm}
            \caption[]%
            {\centering{\small MOS:6.55, CN, 3, score:(7.24 5.88 5.63  {\color{blue}{7.08}})}}     
            \label{fig:qua_2}
        \end{subfigure}
        \hspace{0.7cm}
        \begin{subfigure}[b]{0.2\textwidth}
            \centering 
            \includegraphics[height=5cm]{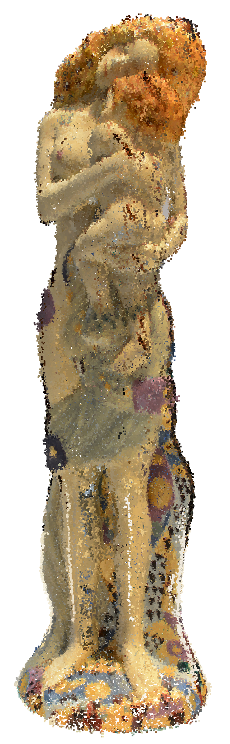}
            \vspace{0.2cm}
            \caption[]%
            {\centering{\small MOS:5.81, DGN, 3, score:(4.86 7.23 6.33  {\color{blue}{6.15}})}}    
            \label{fig:qua_3}
        \end{subfigure}
        \hspace{0.7cm}
        \begin{subfigure}[b]{0.2\textwidth} 
            \centering 
            \includegraphics[height=5cm]{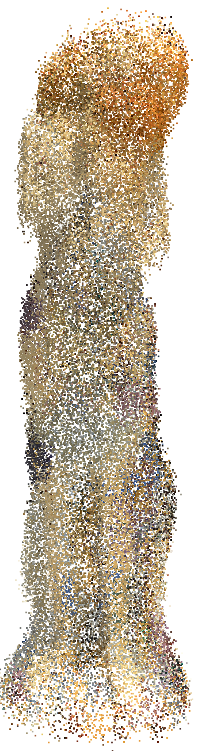}
            \vspace{0.2cm}
            \caption[]%
            {\centering{\small MOS:1.00, DGN, 6, score:(1.85 2.45 2.34 {\color{blue}{1.67}})}}   
            \label{fig:qua_4}
        \end{subfigure}
        
        \begin{subfigure}[b]{0.2\textwidth}
            \centering
            \includegraphics[height=5cm]{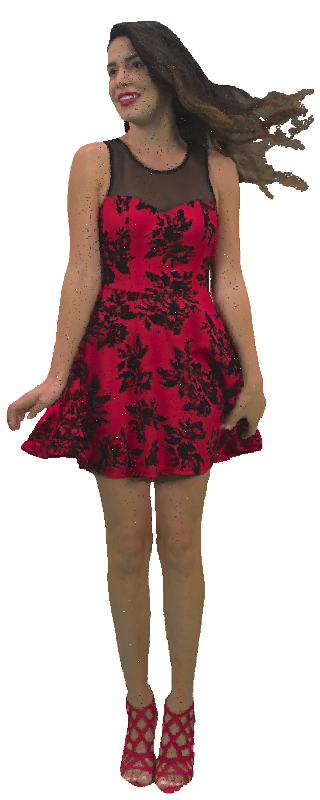}
            \vspace{0.2cm}
            \caption[]%
            {\centering{\small MOS:6.80, DS, 2, score:(5.38 5.49 6.32  {\color{blue}{6.97}})}}    
            \label{fig:qua_5}
        \end{subfigure}
        \hspace{0.7cm}
        \begin{subfigure}[b]{0.2\textwidth}  
            \centering 
            \includegraphics[height=5cm]{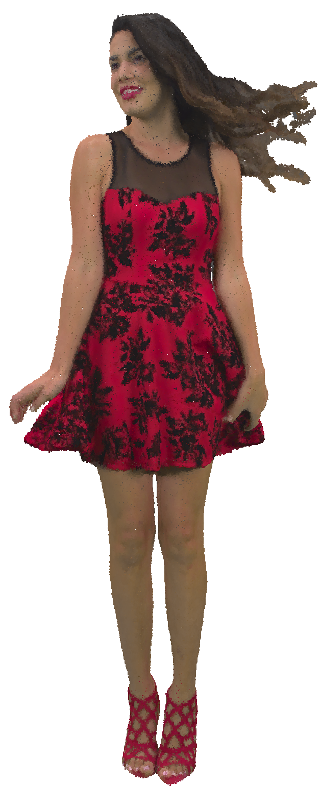}
            \vspace{0.2cm}
            \caption[]%
            {\centering{\small MOS:7.35, DGN, 2, score:(5.86 5.52 6.74  {\color{blue}{7.39}})}}     
            \label{fig:qua_6}
        \end{subfigure}
        \hspace{0.7cm}
        \begin{subfigure}[b]{0.2\textwidth}
            \centering 
            \includegraphics[height=5cm]{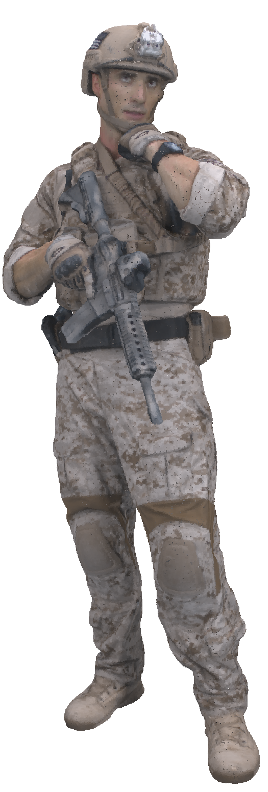}
            \vspace{0.2cm}
            \caption[]%
            {\centering{\small MOS:8.61, DS, 2, score:(6.34 5.64 7.28  {\color{blue}{7.86}})}}    
            \label{fig:qua_7}
        \end{subfigure}
        \hspace{0.7cm}
        \begin{subfigure}[b]{0.2\textwidth} 
            \centering 
            \includegraphics[height=5cm]{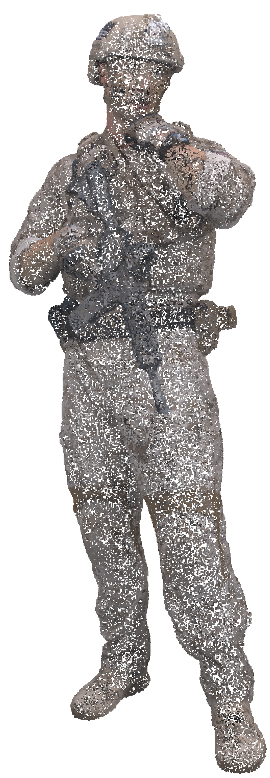}
            \vspace{0.2cm}
            \caption[]%
            {\centering{\small MOS:4.11, DS, 5, score:(4.54 5.07 2.79 {\color{blue}{4.18}})}}   
            \label{fig:qua_8}
        \end{subfigure}
        \caption
        {Qualitative analysis of PCQA on SJTU-PCQA database. In each subfigure, ground truth MOS, the distortion type (DGN represents downsampling and gaussian noise. CN, DS represents color noise and downsampling, respectively) the distortion degree (1-6) and the predicted score of the compared metrics are presented. The predicted score follows the order: PQA-Net, IT-PCQA, ResSCNN, GPA-Net(MT).} 
        \label{fig:qualitative}
    \end{figure*}
    
\subsection{Implementation Details} \label{sec:experiments}
The proposed GPA-Net is implemented on the PyTorch \cite{paszke2019pytorch}. The ADAM optimization algorithm is employed to optimize the model. The learning rate, momentum and drop rate are set to 0.0003, 0.9 and 0.5, respectively.
The batch size is set to 4 through gradient accumulation to reduce the impact of noise and address the variance of the number of input points. 
The $\lambda_1, \lambda_2$ and $\lambda_3$ are tuned automatically in DWA strategy.
The support number in a GPAConv kernel is set to 4. 
The number of keypoints, neighbors and the dimension of feature in the 5 GPAConv layers in Figure \ref{fig:arch} are set to (25000, 5000, 2048, 512, 32), (100, 50, 40, 15, 10) and (16, 64, 128, 256, 512), respectively. The dimension of the hidden layers of the full-connected layers is 64.

To compare with other metrics fairly and reduce the content bias, cross validation of 10 splits is employed for SJTU-PCQA and WPC because of the limited dataset scale. Specifically, for SJTU-PCQA, 2 groups of point cloud are chosen randomly as the testing set and the remaining 7 groups of point cloud are taken as the training set. This procedure is repeated 10 times to get the average results and similar strategy is employed for WPC where 4 groups of point cloud are chosen as the testing set. As for LS-PCQA with a large dataset scale, train-val-test strategy is employed, where we set 80\% for training, 10\% for validation and 10\% for testing. Similarly, this procedure is repeated 5 times to avoid the content bias and the average performance is recorded as the final results. Note that there is no content overlap among the training, validation and testing set.

To ensure the consistency between the predicted scores and MOSs, nonlinear Logistic-4 regression is used to map the predicted scores to the same dynamic range following the recommendations suggested by the Video Quality Experts Group (VQEG) \cite{antkowiak2000final, sheikh2006image}.
This nonlinear regression is monotonic, and all SROCC is calculated before the mapping.
The employed quality assessment evaluation indexes are Spearman rank order correlation coefficient (SROCC) for prediction monotonicity, and Pearson linear correlation coefficient (PLCC) for prediction accuracy. The larger SROCC or PLCC comes with better performance.

\subsection{Comparison of Prediction Performance}
To investigate the prediction performance of GPA-Net comprehensively, 6 full-reference metrics, including MSE-P2point (M-p2po), MSE-P2plane (M-p2pl) \cite{mekuria2016evaluation,tian2017geometric}, Hausdorff-P2point (H-p2po), Hausdorff-P2plane (H-p2pl), PCQM \cite{meynet2020pcqm}, GraphSIM \cite{yang2020inferring} and 3 state-of-the-art no-reference metrics, including ResSCNN, IT-PCQA and PQA-Net, are used for comparison.

Firstly, to compare the metrics in general, we present the prediction performance of the metrics on the SJTU-PCQA, LS-PCQA and WPC dataset. The results are shown in Table \ref{tab:sjtu} - \ref{tab:wpc}. We can see that GPA-Net with multi-tasks, named GPA-Net(MT), presents better SROCC and PLCC than other no-reference metrics, and on LS-PCQA, the performance of our model is even better than all full-reference metrics. 
In addition, compared with PQA-Net \cite{RN52} in which only point clouds with individual distortion type is considered, the poor performance of it may derive from the superimposed distortions. Finally, IT-PCQA does not perform well because the IT-PCQA is based on domain adaptation, in which the labels of point clouds are not utilized in the training. 

Secondly, to demonstrate that our model can attentively capture the critical features related to distortions, we present a qualitative analysis in the testing set of two splits to compare our model with no-reference metrics intuitively. From Figure \ref{fig:qualitative}, we can see that: i) In all, distortion type and degree affect the perceptual quality significantly, see Figure \ref{fig:qua_1} and \ref{fig:qua_3}, Figure \ref{fig:qua_1} and \ref{fig:qua_2}, Figure \ref{fig:qua_3} and \ref{fig:qua_4}, respectively. ii) The other no-reference metrics are not sensitive to distortion pattern changes in Figures \ref{fig:qua_1}, \ref{fig:qua_2} and Figures \ref{fig:qua_3}, \ref{fig:qua_4} because their predicted scores are relatively close, which is similar for Figures \ref{fig:qua_5},\ref{fig:qua_6} and Figures \ref{fig:qua_7}, \ref{fig:qua_8} iii) For GPA-Net, Figures \ref{fig:qua_1} and \ref{fig:qua_3} can be distinguished, and the predicted scores are relatively precise, which means GPA-Net can capture the intrinsic characteristics of point clouds. Besides, with the multi-task learning framework, GPA-Net 
are sensitive to distortion-related features since the predicted scores vary flexibly when distortion pattern changes, which is critical to the PCQA task. To avoid bias and get a more accurate conclusion, we compare the average score difference across all the samples in the above testing set. The results are in Table \ref{tab:difference}, which show that the variation degree of the predicted scores of GPA-Net is closer to the ground truth than other metrics.

\begin{table}[h]
    \centering
    \caption{Comparison of average difference of predicted scores of no-reference metrics when distortion pattern changes.}
    \begin{tabular}{ccc}
    \toprule
         & type changes & degree increases by 1 \\
         \midrule
        MOS difference & \textbf{0.86} & \textbf{0.52} \\
        ResSCNN & 0.41 & 0.36  \\
        PQA-Net & 0.47 & 0.38 \\
        IT-PCQA & 0.45 & 0.35 \\
        GPA-Net & \textbf{0.67} & \textbf{0.46}\\
    \bottomrule
    \end{tabular}
    \label{tab:difference}
\end{table}

\begin{table*}[h]
    \centering
    \caption{Comparison of generalization performance of no-reference methods in cross-dataset validation}
    \begin{tabular}{c|cc|cc|cc|cc|cc|cc}
    \toprule
        Train on &  \multicolumn{4}{c}{LS-PCQA} & \multicolumn{4}{c}{SJTU-PCQA} & \multicolumn{4}{c}{WPC}\\
    \midrule
        Test on &  \multicolumn{2}{c}{SJTU-PCQA} & \multicolumn{2}{|c}{WPC} & \multicolumn{2}{|c}{LS-PCQA} & \multicolumn{2}{|c}{WPC} & \multicolumn{2}{|c}{LS-PCQA} & \multicolumn{2}{|c}{SJTU-PCQA} \\
    \midrule
         & SROCC & \multicolumn{1}{c|}{PLCC}  & SROCC & \multicolumn{1}{c|}{PLCC} & SROCC & \multicolumn{1}{c|}{PLCC} & SROCC & \multicolumn{1}{c|}{PLCC} & SROCC & \multicolumn{1}{c|}{PLCC} & SROCC & \multicolumn{1}{c|}{PLCC}  \\
    \midrule
         ResSCNN & 0.532 & 0.546 & 0.478 & 0.466 & \textbf{0.327} & \textbf{0.342} & 0.258 & 0.269 & \textbf{0.325} & 0.336 & \textbf{0.535} & 0.572 \\
        IT-PCQA & 0.512 & 0.533 & 0.488 & 0.499 & 0.265 & 0.258 & 0.410 & 0.429 & 0.264 & 0.285 & 0.482 & 0.471 \\
        PQA-Net & 0.365 & 0.383 & 0.411 & 0.427 & 0.277 & 0.271 & 0.265 & 0.275 & 0.234 & 0.253 & 0.477 & 0.486 \\
        GPA-Net & \textbf{0.582} & \textbf{0.598} & \textbf{0.492} & \textbf{0.510} & 0.323 & 0.336 & \textbf{0.424} & \textbf{0.431} & 0.312 & \textbf{0.343} & \textbf{0.535} & \textbf{0.574} \\
    \bottomrule
    \end{tabular}
    \label{tab:crossdataset}
\end{table*}

\begin{table*}[h]
    \centering
    \caption{ \centering{Comparison of robustness of no-reference metrics on SJTU-PCQA database. "+0.2" means translation by 0.2 \hspace{\textwidth} times the length along $x$ axis. "*2" means doubling the scale. "$45^\circ$" means rotation by 45 degrees around $x$ axis.}}
    \begin{tabular}{cccccccccc}
    \toprule
        &\multicolumn{4}{c}{SROCC} & & \multicolumn{4}{c}{PLCC}\\
        \midrule
        & ResSCNN & IT-PCQA & PQA-Net & GPA-Net(MT) &
        & ResSCNN & IT-PCQA & PQA-Net & GPA-Net(MT)\\
        \midrule
        - & 0.834 & 0.597 & 0.689 & \textbf{0.875} &
        - & 0.863 & 0.618 & 0.705 & \textbf{0.886}\\
        +0.2 & 0.812 & 0.570 & 0.680 & \textbf{0.875} &
        +0.2 & 0.840 & 0.592 & 0.697 & \textbf{0.886}\\
        *2 & 0.796 & 0.563 & 0.677 & \textbf{0.875} &
        *2 & 0.812 & 0.585 & 0.689 & \textbf{0.886}\\
        $45^\circ$ & 0.788 & 0.496 & 0.626 & \textbf{0.875} &
        $45^\circ$ & 0.794 & 0.527 & 0.635 & \textbf{0.886}\\
    \bottomrule
    \end{tabular}
    \label{tab:robust}
\end{table*}

In summary, experiments show that our GPA-Net presents a competitive performance comprehensively, compared to other state-of-the-art no-reference PCQA metrics. Moreover, qualitative analysis demonstrate that our multi-task model is attentive to distortion properties and achieves superior performance than other no-reference metrics.

\subsection{Comparison of Generalization Performance}
In this section, we first evaluate the generalization performance of the no-reference metrics across different datasets, then we evaluate the generalization performance for unknown distortions, including Leave-One-Distortion-Out and Leave-One-Degree-Out evaluations.

To evaluate cross-dataset performance of the no-reference metrics, we use 90\% of LS-PCQA for training and the remaining 10\% for validation, and test the trained models on the complete SJTU-PCQA and WPC. When using SJTU-PCQA or WPC for training, we select the best model by the minimal loss and test the trained model on the complete LS-PCQA. The cross-dataset evaluation results are as in Table \ref{tab:crossdataset}:

\begin{table}[h]
    \centering
    \caption{Comparison of generalization performance of no-reference metrics in Leave-One-Distortion-Out cross evaluation.}
    \begin{tabular}{ccccc}
    \toprule
         & ResSCNN & IT-PCQA & PQA-Net & GPA-Net(MT)\\
         \midrule
        SROCC & 0.754 & 0.588 & 0.542 & \textbf{0.782}\\
        PLCC & 0.782 & 0.603 & 0.591 & \textbf{0.807}\\
    \bottomrule
    \end{tabular}
    \label{tab:crosstype}
\end{table}

\begin{table}[h]
    \centering
    \caption{Comparison of generalization performance of no-reference metrics in Leave-One-Degree-Out cross evaluation.}
    \begin{tabular}{ccccc}
    \toprule
         & ResSCNN & IT-PCQA & PQA-Net & GPA-Net(MT)\\
         \midrule
        SROCC & 0.762 & 0.643 & 0.578 & \textbf{0.784}\\
        PLCC & 0.785 & 0.664 & 0.609 & \textbf{0.803}\\
    \bottomrule
    \end{tabular}
    \label{tab:crossdegree}
\end{table}
From Table \ref{tab:crossdataset} we can see that the cross-dataset performances of these no-reference metrics are limited. One reason is that the distortion types considered in these datasets vary greatly. For example, in SJTU-PCQA, most of the distortions are geometric distortion (e.g., gaussian noise, octree compression noise); while in LS-PCQA, the major are attribute distortions (e.g., color noise, contrast noise). The differences of distortions make it extremely difficult for these learning-based no-reference metrics to learn general features. However, despite the great distortion differences, our GPA-Net still presents competitive performance.

To further evaluate the generalization performances for unknown distortion patterns, we conduct Leave-One-Distortion-Out and Leave-One-Degree-Out evaluations on SJTU-PCQA dataset. In implementation, we use 6 kinds of distortion types for training and the remaining 1 kind of distortion type for testing. The strategy is similar for different distortion degrees. The test results listed in Table \ref{tab:crosstype} and \ref{tab:crossdegree}. We can see that the performance decline is less than cross-dataset evaluation, and our GPA-Net still performs the best among the no-reference metrics.

In conclusion, experiments show that the proposed GPA-Net presents the best generalization performance compared to other state-of-the-art no-reference PCQA metrics.

\subsection{Comparison of Model Robustness}
In this section, we compare the robustness of our GPA-Net with other no-reference metrics in terms of sample translation, scaling and rotation. Specifically, we translate the point clouds in the testing set by 0.2 times the length along the global $x$ axis, double the scale, and rotate the point clouds by 45 degrees around $x$ axis, respectively. Note that we neglect the data augmentations in training for all the compared metrics, including GPA-Net.

In Table \ref{tab:robust}, we present the performance of no-reference metrics under translation, scaling and rotation. We can see that our GPA-Net is totally invariant to the above 3 perturbations, while other metrics present noticeable  performance decline, especially under the rotation with $45^\circ$ around global $x$ axis.

\subsection{Comparison of Computation Complexity and Inference Time}
In this section, we compare the computation complexity and inference time in Table \ref{tab:computation} including the number of model parameters and inference time. Out of fairness, the preprocessing operations (voxelization for ResSCNN, projection and rendering for IT-PCQA and PQA-Net) are taken into consideration for inference time measurement. Note that the rendering process for 2D-based methods is based on PyTorch3D \cite{ravi2020accelerating}, using GPU for acceleration. The inference time is the average result measured upon selected samples from SJTU-PCQA on a NVIDIA RTX3090 GPU.

\begin{table}[h]
    \centering
    \caption{The number of parameters and inference time of different no-reference methods }
    \begin{tabular}{c|cccc}
    \toprule
        Method & IT-PCQA & PQA-Net & ResSCNN & GPA-Net \\
    \midrule
        \#Parameters &  0.61M & \textbf{0.22M} & 1.23M &1.58M\\
        Inference time  &  2.87s & 4.86s & 1.92s & \textbf{0.89s}\\
    \bottomrule
    \end{tabular}
    \label{tab:computation}
\end{table}

From Table \ref{tab:computation}, we can conclude that:
\begin{itemize}
    \item The 3D-based models (ResSCNN and GPA-Net) are larger in model size compared 2D-based methods (IT-PCQA and PQA-Net) but faster in terms of inference time. This is because projection and rendering for 2D-based methods are relatively time-consuming.
    \item Although GPA-Net is the largest for model size, but faster than other methods by a large margin, which can prove the efficiency of GPA-Net.
\end{itemize}
\subsection{Ablation Study}
In this section, we first compare GPA-Net with single-task, i.e., GPA-Net(ST), with other point-based or graph-based feature encoders. Next, we evaluate the performance of GPA-Net with different $\lambda_1$ and $\lambda_2$. Then, we examine the effectiveness of the coordinate normalization. At last, we compare the performance with different reorientations of SLRF.
\begin{figure}[t]
    \centering
    \includegraphics[width=0.48\textwidth]{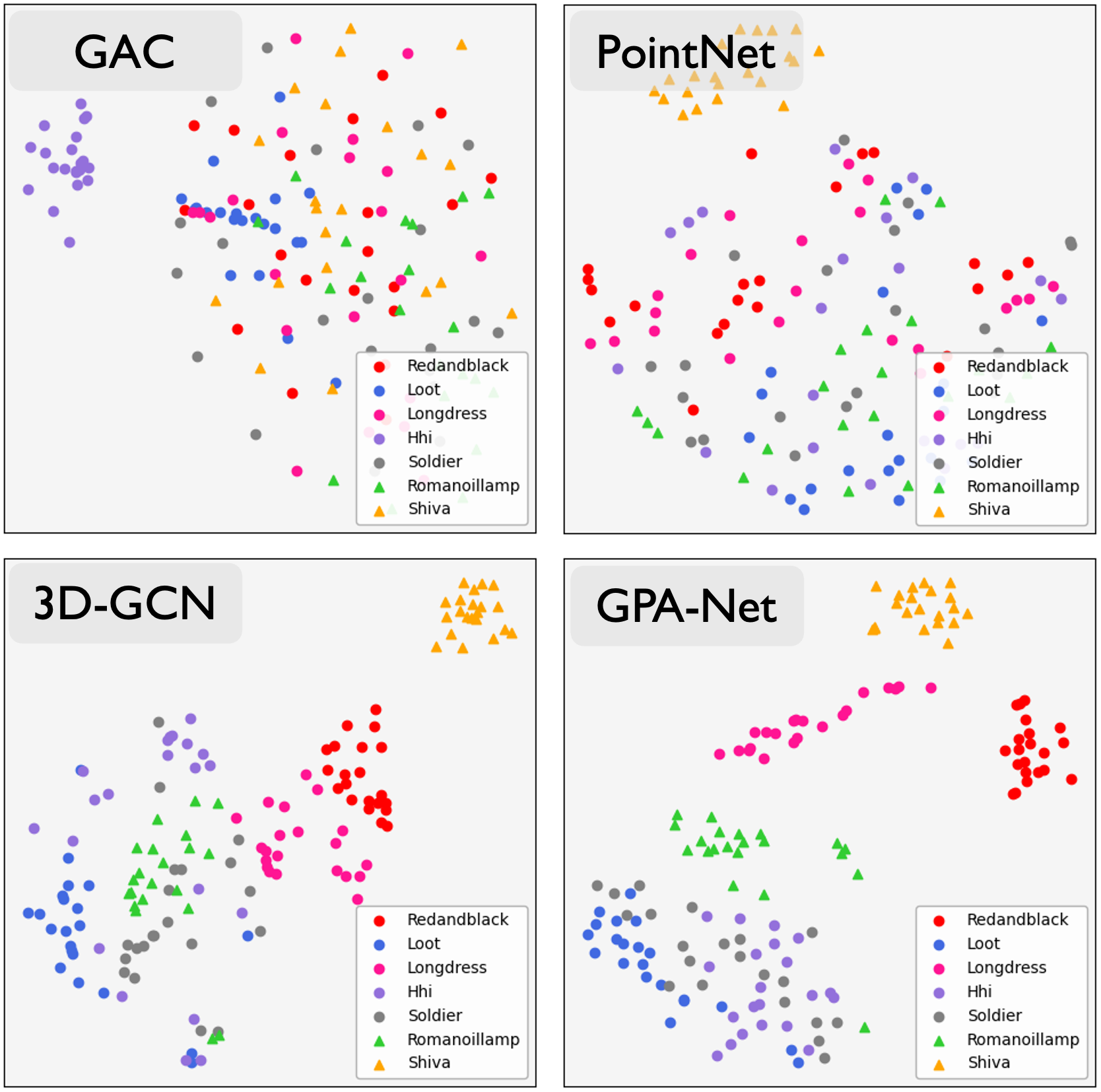}
    \caption{T-SNE 2D embedding of the feature space of the compared encoders on SJTU-PCQA under distortion degree 1-3. The scattered points are color encoded depending on the content of point clouds.}
    \label{fig:tsne}
\end{figure}
\begin{table}[t]
    \centering
    \caption{Performance of different feature encoders.}
    \begin{tabular}{ccccc}
    \toprule
         & PointNet & GAC & 3D-GCN & GPA-Net(ST)\\
         \midrule
        SROCC & 0.423 & 0.723 & 0.824 & \textbf{0.847}\\
        PLCC & 0.472 & 0.674 & 0.835 & \textbf{0.856}\\
    \bottomrule
    \end{tabular}
    \label{tab:encoder}
\end{table}

\subsubsection{Comparison of feature encoders.}
In Table \ref{tab:encoder}, we present the performance of different feature encoders, including PointNet \cite{qi2017pointnet}, GAC \cite{wang2019graph}, 3D-GCN \cite{lin2020convolution} and the proposed GPA encoder. 3D-GCN performs relatively well because geometric information is explicitly utilized when computing cosine similarities. However, referring to Figure \ref{fig:toy}, 3D-GCN may make mistakes while only using partial graph information, leading to poorer performance than GPA-Net(ST). The GPA encoder takes all the patch points into consideration and computes both direction similarities and length similarities, which contribute to the best performance for no-reference PCQA.

In Figure \ref{fig:tsne}, we further investigate the expressiveness of the feature encoders for dense point clouds by performing t-SNE algorithm \cite{van2008visualizing} on the high-dimensional feature after max-pooling of each trained encoder in the training set of SJTU-PCQA. T-SNE reduces the feature dimensions for visualization and preserve the significant structure of the high-dimensional features. We can see from Figure \ref{fig:tsne} that all the encoders except GAC cannot distinguish $Loot$, $Hhi$ and $Soldier$ because these 3 point clouds are close in the content since they are all males and their texture changes smoothly, which are contrary to $Redandblack$ and $Longdress$. We can also see that compared to other encoders, our model not only achieves good clustering results, but also let point clouds with relatively significant difference in content be further separated in the feature space, which shows the potential to perform well when more point clouds with different content or distortions included.



\subsubsection{Comparison of performance without rotation invariance.}
In Table \ref{tab:ab_robust}, we present the SROCC and PLCC of GPA-Net when the rotation invariance is removed. We can see that when GPA-Net is rotation variant, the prediction performance is better because the original positions of the patches are preserved before feature extraction. However, when the point clouds rotate around $x, y$ or $z$ axis, the performance degrades obviously, and the degradation increase rapidly with the rotation degree growing. The performance is the worst when the point clouds rotate around $y$ axis.
\begin{table}[h]
    \centering
    \caption{{Performance of GPA-Net with different local coordinate normalization. GPA-Net(V) is rotation variant. $x, y, z$ mean the axis around with point clouds rotate.}}
    
    \begin{tabular}{ccccc}
    \toprule
        &\multicolumn{2}{c}{SROCC}  & \multicolumn{2}{c}{PLCC}\\
        \midrule
        & GPA-Net(V) & GPA-Net 
        & GPA-Net(V) & GPA-Net\\
        \midrule
        - & \textbf{0.886} & 0.875 &
         \textbf{0.897} & 0.886\\
        $x,45^\circ$ & 0.853 &  \textbf{0.875} &
         0.870 &  \textbf{0.886}\\
        $x,90^\circ$ & 0.845 &  \textbf{0.875} &
         0.865 &  \textbf{0.886}\\
        $x,180^\circ$ & 0.842 &  \textbf{0.875} &
         0.860 &  \textbf{0.886}\\
        $y,45^\circ$ & 0.844 &  \textbf{0.875} &
         0.863 &  \textbf{0.886}\\
         $y,90^\circ$ & 0.845 &  \textbf{0.875} &
         0.859 &  \textbf{0.886}\\
         $y,180^\circ$ & 0.837 &  \textbf{0.875} &
         0.867 &  \textbf{0.886}\\
        $z,45^\circ$ & 0.861 & \textbf{0.875} &
         0.869 & \textbf{0.886}\\
         $z,90^\circ$ & 0.859 &  \textbf{0.875} &
         0.868 &  \textbf{0.886}\\
         $z,180^\circ$ & 0.855 &  \textbf{0.875} &
         0.865 &  \textbf{0.886}\\
    \bottomrule
    \end{tabular}
    \label{tab:ab_robust}
\end{table}
\begin{table}[h]
    \centering
    \caption{Performance of GPA-Net with different reorientations. Reorientations of local $z, x$ axes are in the bracket.}
    \begin{tabular}{ccccccc}
    \toprule
         & $(x,y)$ & $(x,z)$ & $(y,x)$ & $(z,x)$ & $(z,y)$ & $(y,z)$\\
         \midrule
        SROCC & 0.842 & 0.872 & 0.843  & 0.836 & 0.849 & \textbf{0.875}\\
        PLCC & 0.848 & 0.876 & 0.859  & 0.847 & 0.860 & \textbf{0.886}\\
    \bottomrule
    \end{tabular}
    \label{tab:global_axis}
\end{table}

\subsubsection{Comparison of global axes that LRF reorients to.}\label{sec:ab_lrf}
In Table \ref{tab:global_axis}, we present the SROCC and PLCC of GPA-Net when the local axes reorient to different global axes. We can see that GPA-Net performs the best when the local $z, x$ axes reorient to global $y, z$ axes.

\subsubsection{Result variation depending on parameter setting in the first GPA layer}
\begin{figure}[h]
    \centering
    \includegraphics[width=0.5\textwidth]{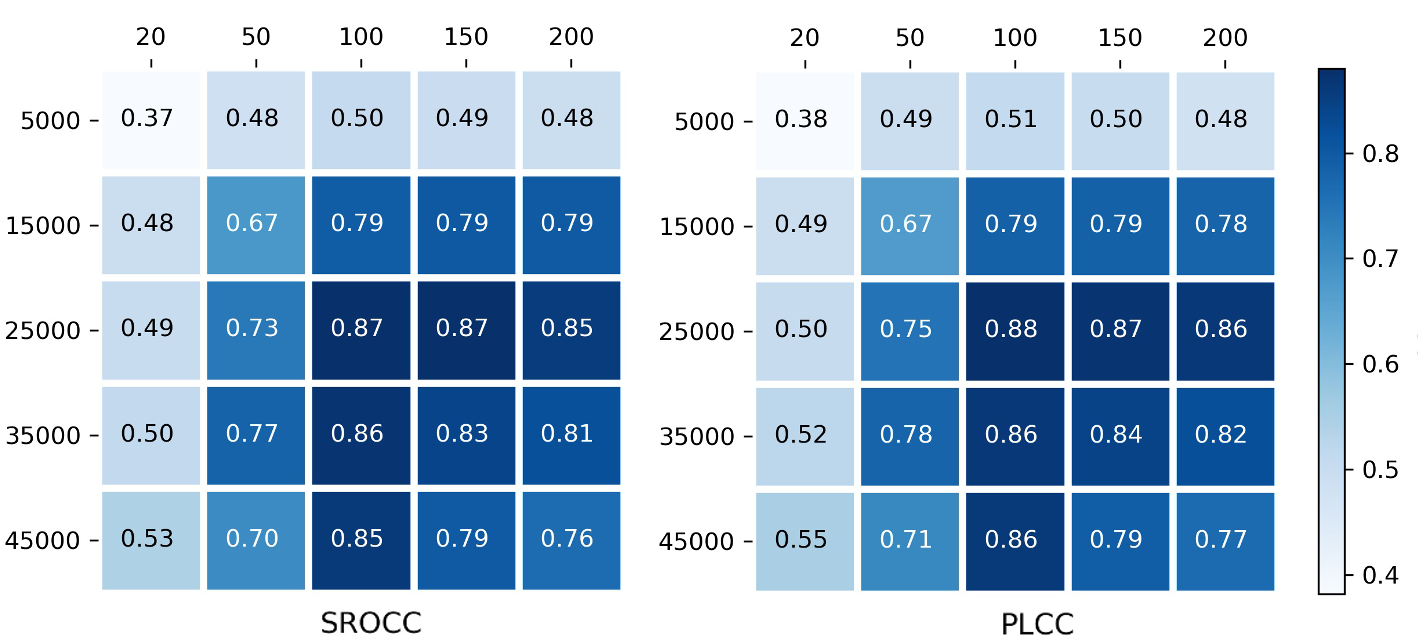}
    \caption{Result variations depending on parameter setting in the first GPA layer}
    \label{fig:heatmap}
\end{figure}

In Figure \ref{fig:heatmap}, we present the SROCC and PLCC depending on the parameter setting including the number of keypoints and neighbor in the first GPA layer, which have the largest effect on the results among the selected parameters.

From Figure \ref{fig:heatmap} we can see that the GPA-Net performs the best when the numbers of keypoints and neighbors are set to 25000 and 100. This is because the point cloud patches of this scale cover the whole point cloud and have slight overlap between them, as described in \cite{lin2020convolution}.

\section{Conclusion}\label{sec:conclusion}
In this paper, we propose a novel no-reference PCQA metric (GPA-Net) using a multi-task graph convolutional network based on the proposed graph convolution kernel (GPAConv), which can attentively extract perturbation of structure and texture. Besides, to achieve the shift, scale and rotation invariance, we propose a coordinate normalization module. Experiments show that 
our model outperforms the current no-reference PCQA metrics in two independent point cloud databases.

\ifCLASSOPTIONcompsoc
  \section*{Acknowledgments}
\else
  \section*{Acknowledgment}
\fi
This paper is supported in part by National Key R\&D Program of China (2018YFE0206700), National Natural Science Foundation of China (61971282, U20A20185). The corresponding author is Yiling Xu(e-mail: yl.xu@sjtu.edu.cn).

\ifCLASSOPTIONcaptionsoff
  \newpage
\fi




\bibliographystyle{IEEEtran}

\begin{thebibliography}{10}
\providecommand{\url}[1]{#1}
\csname url@samestyle\endcsname
\providecommand{\newblock}{\relax}
\providecommand{\bibinfo}[2]{#2}
\providecommand{\BIBentrySTDinterwordspacing}{\spaceskip=0pt\relax}
\providecommand{\BIBentryALTinterwordstretchfactor}{4}
\providecommand{\BIBentryALTinterwordspacing}{\spaceskip=\fontdimen2\font plus
\BIBentryALTinterwordstretchfactor\fontdimen3\font minus
  \fontdimen4\font\relax}
\providecommand{\BIBforeignlanguage}[2]{{%
\expandafter\ifx\csname l@#1\endcsname\relax
\typeout{** WARNING: IEEEtran.bst: No hyphenation pattern has been}%
\typeout{** loaded for the language `#1'. Using the pattern for}%
\typeout{** the default language instead.}%
\else
\language=\csname l@#1\endcsname
\fi
#2}}
\providecommand{\BIBdecl}{\relax}
\BIBdecl

\bibitem{alexiou2017towards}
E.~Alexiou, E.~Upenik, and T.~Ebrahimi, ``Towards subjective quality assessment
  of point cloud imaging in augmented reality,'' in \emph{2017 IEEE 19th
  International Workshop on Multimedia Signal Processing}, 2017, pp. 1--6.

\bibitem{schwarz2018emerging}
S.~Schwarz, M.~Preda, V.~Baroncini, M.~Budagavi, P.~Cesar, P.~A. Chou, R.~A.
  Cohen, M.~Krivoku{\'c}a, S.~Lasserre, Z.~Li \emph{et~al.}, ``Emerging mpeg
  standards for point cloud compression,'' \emph{IEEE Journal on Emerging and
  Selected Topics in Circuits and Systems}, vol.~9, no.~1, pp. 133--148, 2018.

\bibitem{zhao2020quality}
H.~Zhao, Q.~Yang, Y.~Xu \emph{et~al.}, ``A quality metric for 3d lidar point
  cloud based on vision tasks,'' in \emph{2020 IEEE International Symposium on
  Broadband Multimedia Systems and Broadcasting}, 2020, pp. 1--5.

\bibitem{bletterer2020local}
A.~Bletterer, F.~Payan, and M.~Antonini, ``A local graph-based structure for
  processing gigantic aggregated 3d point clouds,'' \emph{IEEE Transactions on
  Visualization and Computer Graphics}, 2020.

\bibitem{liu2020model}
Q.~Liu, H.~Yuan, J.~Hou, R.~Hamzaoui, and H.~Su, ``Model-based joint bit
  allocation between geometry and color for video-based 3d point cloud
  compression,'' \emph{IEEE Transactions on Multimedia}, vol.~23, pp.
  3278--3291, 2020.

\bibitem{su2019perceptual}
H.~Su, Z.~Duanmu, W.~Liu, Q.~Liu, and Z.~Wang, ``Perceptual quality assessment
  of 3d point clouds,'' in \emph{2019 IEEE International Conference on Image
  Processing}, 2019, pp. 3182--3186.

\bibitem{mekuria2016evaluation}
R.~Mekuria, Z.~Li, C.~Tulvan, and P.~Chou, ``Evaluation criteria for point
  cloud compression,'' \emph{ISO/IEC MPEG}, no. 16332, 2016.

\bibitem{tian2017geometric}
D.~Tian, H.~Ochimizu, C.~Feng, R.~Cohen, and A.~Vetro, ``Geometric distortion
  metrics for point cloud compression,'' in \emph{2017 IEEE International
  Conference on Image Processing}, 2017, pp. 3460--3464.

\bibitem{alexiou2018point}
E.~Alexiou and T.~Ebrahimi, ``Point cloud quality assessment metric based on
  angular similarity,'' in \emph{2018 IEEE International Conference on
  Multimedia and Expo}, 2018, pp. 1--6.

\bibitem{meynet2020pcqm}
G.~Meynet, Y.~Nehm{\'e}, J.~Digne, and G.~Lavou{\'e}, ``Pcqm: A full-reference
  quality metric for colored 3d point clouds,'' in \emph{2020 Twelfth
  International Conference on Quality of Multimedia Experience}, 2020, pp.
  1--6.

\bibitem{yang2020inferring}
Q.~Yang, Z.~Ma, Y.~Xu, Z.~Li, and J.~Sun, ``Inferring point cloud quality via
  graph similarity,'' \emph{IEEE Transactions on Pattern Analysis and Machine
  Intelligence}, 2020.

\bibitem{zhang2021ms}
Y.~Zhang, Q.~Yang, and Y.~Xu, ``Ms-graphsim: Inferring point cloud quality via
  multiscale graph similarity,'' in \emph{Proceedings of the 29th ACM
  International Conference on Multimedia}, 2021, pp. 1230--1238.

\bibitem{yang2022mped}
Q.~Yang, Y.~Zhang, S.~Chen, Y.~Xu, J.~Sun, and Z.~Ma, ``Mped: Quantifying point
  cloud distortion based on multiscale potential energy discrepancy,''
  \emph{IEEE Transactions on Pattern Analysis and Machine Intelligence}, 2022.

\bibitem{liu2022point}
Y.~Liu, Q.~Yang, Y.~Xu, and L.~Yang, ``Point cloud quality assessment: Dataset
  construction and learning-based no-reference metric,'' \emph{ACM Transactions
  on Multimedia Computing, Communications, and Applications}, 2022.

\bibitem{RN52}
Q.~Liu, H.~Yuan, H.~Su, H.~Liu, Y.~Wang, H.~Yang, and J.~Hou, ``Pqa-net: Deep
  no reference point cloud quality assessment via multi-view projection,''
  \emph{IEEE Transactions on Circuits and Systems for Video Technology}, pp.
  1--1, 2021.

\bibitem{yang2022no}
Q.~Yang, Y.~Liu, S.~Chen, Y.~Xu, and J.~Sun, ``No-reference point cloud quality
  assessment via domain adaptation,'' in \emph{Proceedings of the IEEE/CVF
  Conference on Computer Vision and Pattern Recognition}, 2022, pp.
  21\,179--21\,188.

\bibitem{liu2019relation}
Y.~Liu, B.~Fan, S.~Xiang, and C.~Pan, ``Relation-shape convolutional neural
  network for point cloud analysis,'' in \emph{Proceedings of the IEEE/CVF
  Conference on Computer Vision and Pattern Recognition}, 2019, pp. 8895--8904.

\bibitem{watson1997digital}
A.~B. Watson and C.~H. Null, ``Digital images and human vision,'' in
  \emph{Electronic Imaging Science and Technology Conference}, 1997.

\bibitem{yang2020predicting}
Q.~Yang, H.~Chen, Z.~Ma, Y.~Xu, R.~Tang, and J.~Sun, ``Predicting the
  perceptual quality of point cloud: A 3d-to-2d projection-based exploration,''
  \emph{IEEE Transactions on Multimedia}, vol.~23, pp. 3877--3891, 2020.

\bibitem{liu2021reduced}
Q.~Liu, H.~Yuan, R.~Hamzaoui, H.~Su, J.~Hou, and H.~Yang, ``Reduced reference
  perceptual quality model with application to rate control for video-based
  point cloud compression,'' \emph{IEEE Transactions on Image Processing},
  vol.~30, pp. 6623--6636, 2021.

\bibitem{perry2020quality}
S.~Perry, H.~P. Cong, L.~A. da~Silva~Cruz, J.~Prazeres, M.~Pereira,
  A.~Pinheiro, E.~Dumic, E.~Alexiou, and T.~Ebrahimi, ``Quality evaluation of
  static point clouds encoded using mpeg codecs,'' in \emph{2020 IEEE
  International Conference on Image Processing}, 2020, pp. 3428--3432.

\bibitem{lin2020convolution}
Z.-H. Lin, S.-Y. Huang, and Y.-C.~F. Wang, ``Convolution in the cloud: Learning
  deformable kernels in 3d graph convolution networks for point cloud
  analysis,'' in \emph{Proceedings of the IEEE/CVF Conference on Computer
  Vision and Pattern Recognition}, 2020, pp. 1800--1809.

\bibitem{zhang2022no}
Z.~Zhang, W.~Sun, X.~Min, T.~Wang, W.~Lu, and G.~Zhai, ``No-reference quality
  assessment for 3d colored point cloud and mesh models,'' \emph{IEEE
  Transactions on Circuits and Systems for Video Technology}, 2022.

\bibitem{liu2019end}
S.~Liu, E.~Johns, and A.~J. Davison, ``End-to-end multi-task learning with
  attention,'' in \emph{Proceedings of the IEEE/CVF conference on computer
  vision and pattern recognition}, 2019, pp. 1871--1880.

\bibitem{chetouani2021deep}
A.~Chetouani, M.~Quach, G.~Valenzise, and F.~Dufaux, ``Deep learning-based
  quality assessment of 3d point clouds without reference,'' in \emph{2021 IEEE
  International Conference on Multimedia \& Expo Workshops}, 2021, pp. 1--6.

\bibitem{simonyan2014very}
K.~Simonyan and A.~Zisserman, ``Very deep convolutional networks for
  large-scale image recognition,'' \emph{arXiv preprint arXiv:1409.1556}, 2014.

\bibitem{fan2022no}
Y.~Fan, Z.~Zhang, W.~Sun, X.~Min, W.~Lu, T.~Wang, N.~Liu, and G.~Zhai, ``A
  no-reference quality assessment metric for point cloud based on captured
  video sequences,'' \emph{arXiv preprint arXiv:2206.05054}, 2022.

\bibitem{tliba2022point}
M.~Tliba, A.~Chetouani, G.~Valenzise, and F.~Dufaux, ``Point cloud quality
  assessment using cross-correlation of deep features,'' in \emph{Proceedings
  of the 2nd Workshop on Quality of Experience in Visual Multimedia
  Applications}, 2022, pp. 63--68.

\bibitem{qi2017pointnet}
C.~R. Qi, H.~Su, K.~Mo, and L.~J. Guibas, ``Pointnet: Deep learning on point
  sets for 3d classification and segmentation,'' in \emph{Proceedings of the
  IEEE/CVF Conference on Computer Vision and Pattern Recognition}, 2017, pp.
  652--660.

\bibitem{qi2017pointnet++}
C.~R. Qi, L.~Yi, H.~Su, and L.~J. Guibas, ``Pointnet++: Deep hierarchical
  feature learning on point sets in a metric space,'' \emph{Advances in Neural
  Information Processing Systems}, vol.~30, 2017.

\bibitem{thomas2019kpconv}
H.~Thomas, C.~R. Qi, J.-E. Deschaud, B.~Marcotegui, F.~Goulette, and L.~J.
  Guibas, ``Kpconv: Flexible and deformable convolution for point clouds,'' in
  \emph{Proceedings of the IEEE/CVF International Conference on Computer
  Vision}, 2019, pp. 6411--6420.

\bibitem{li2021rotation}
X.~Li, R.~Li, G.~Chen, C.-W. Fu, D.~Cohen-Or, and P.-A. Heng, ``A
  rotation-invariant framework for deep point cloud analysis,'' \emph{IEEE
  Transactions on Visualization and Computer Graphics}, 2021.

\bibitem{wang2019dynamic}
Y.~Wang, Y.~Sun, Z.~Liu, S.~E. Sarma, M.~M. Bronstein, and J.~M. Solomon,
  ``Dynamic graph cnn for learning on point clouds,'' \emph{Acm Transactions On
  Graphics}, vol.~38, no.~5, pp. 1--12, 2019.

\bibitem{ravi2020accelerating}
N.~Ravi, J.~Reizenstein, D.~Novotny, T.~Gordon, W.-Y. Lo, J.~Johnson, and
  G.~Gkioxari, ``Accelerating 3d deep learning with pytorch3d,'' \emph{arXiv
  preprint arXiv:2007.08501}, 2020.

\bibitem{wang2019graph}
L.~Wang, Y.~Huang, Y.~Hou, S.~Zhang, and J.~Shan, ``Graph attention convolution
  for point cloud semantic segmentation,'' in \emph{Proceedings of the IEEE/CVF
  Conference on Computer Vision and Pattern Recognition}, 2019, pp.
  10\,296--10\,305.

\bibitem{zhou2021adaptive}
H.~Zhou, Y.~Feng, M.~Fang, M.~Wei, J.~Qin, and T.~Lu, ``Adaptive graph
  convolution for point cloud analysis,'' in \emph{Proceedings of the IEEE/CVF
  International Conference on Computer Vision}, 2021, pp. 4965--4974.

\bibitem{chang2015shapenet}
A.~X. Chang, T.~Funkhouser, L.~Guibas, P.~Hanrahan, Q.~Huang, Z.~Li,
  S.~Savarese, M.~Savva, S.~Song, H.~Su \emph{et~al.}, ``Shapenet: An
  information-rich 3d model repository,'' \emph{arXiv preprint
  arXiv:1512.03012}, 2015.

\bibitem{wu20153d}
Z.~Wu, S.~Song, A.~Khosla, F.~Yu, L.~Zhang, X.~Tang, and J.~Xiao, ``3d
  shapenets: A deep representation for volumetric shapes,'' in
  \emph{Proceedings of the IEEE/CVF Conference on Computer Vision and Pattern
  Recognition}, 2015, pp. 1912--1920.

\bibitem{zhang2021survey}
Y.~Zhang and Q.~Yang, ``A survey on multi-task learning,'' \emph{IEEE
  Transactions on Knowledge and Data Engineering}, 2021.

\bibitem{xu2016multi}
L.~Xu, J.~Li, W.~Lin, Y.~Zhang, L.~Ma, Y.~Fang, and Y.~Yan, ``Multi-task rank
  learning for image quality assessment,'' \emph{IEEE Transactions on Circuits
  and Systems for Video Technology}, vol.~27, no.~9, pp. 1833--1843, 2016.

\bibitem{golestaneh2020no}
S.~A. Golestaneh and K.~Kitani, ``No-reference image quality assessment via
  feature fusion and multi-task learning,'' \emph{arXiv preprint
  arXiv:2006.03783}, 2020.

\bibitem{hou2020content}
J.~Hou, W.~Lin, and B.~Zhao, ``Content-dependency reduction with multi-task
  learning in blind stitched panoramic image quality assessment,'' in
  \emph{2020 IEEE International Conference on Image Processing}, 2020, pp.
  3463--3467.

\bibitem{li2022blind}
A.~Li, J.~Wu, S.~Tian, L.~Li, W.~Dong, and G.~Shi, ``Blind image quality
  assessment based on progressive multi-task learning,'' \emph{Neurocomputing},
  2022.

\bibitem{vaswani2017attention}
A.~Vaswani, N.~Shazeer, N.~Parmar, J.~Uszkoreit, L.~Jones, A.~N. Gomez,
  {\L}.~Kaiser, and I.~Polosukhin, ``Attention is all you need,''
  \emph{Advances in Neural Information Processing Systems}, vol.~30, 2017.

\bibitem{wu2019pointconv}
W.~Wu, Z.~Qi, and L.~Fuxin, ``Pointconv: Deep convolutional networks on 3d
  point clouds,'' in \emph{Proceedings of the IEEE/CVF Conference on Computer
  Vision and Pattern Recognition}, 2019, pp. 9621--9630.

\bibitem{ao2020repeatable}
S.~Ao, Y.~Guo, J.~Tian, Y.~Tian, and D.~Li, ``A repeatable and robust local
  reference frame for 3d surface matching,'' \emph{Pattern Recognition}, vol.
  100, p. 107186, 2020.

\bibitem{bro2008resolving}
R.~Bro, E.~Acar, and T.~G. Kolda, ``Resolving the sign ambiguity in the
  singular value decomposition,'' \emph{Journal of Chemometrics: A Journal of
  the Chemometrics Society}, vol.~22, no.~2, pp. 135--140, 2008.

\bibitem{salti2014shot}
S.~Salti, F.~Tombari, and L.~Di~Stefano, ``Shot: Unique signatures of
  histograms for surface and texture description,'' \emph{Computer Vision and
  Image Understanding}, vol. 125, pp. 251--264, 2014.

\bibitem{kovacs2012rotation}
E.~Kov{\'a}cs, ``Rotation about an arbitrary axis and reflection through an
  arbitrary plane,'' in \emph{Annales Mathematicae et Informaticae}, vol.~40,
  2012, pp. 175--186.

\bibitem{paszke2019pytorch}
A.~Paszke, S.~Gross, F.~Massa, A.~Lerer, J.~Bradbury, G.~Chanan, T.~Killeen,
  Z.~Lin, N.~Gimelshein, L.~Antiga \emph{et~al.}, ``Pytorch: An imperative
  style, high-performance deep learning library,'' \emph{Advances in Neural
  Information Processing Systems}, vol.~32, 2019.

\bibitem{antkowiak2000final}
J.~Antkowiak, T.~Jamal~Baina, F.~V. Baroncini, N.~Chateau, F.~FranceTelecom,
  A.~C.~F. Pessoa, F.~Stephanie~Colonnese, I.~L. Contin, J.~Caviedes, and
  F.~Philips, ``Final report from the video quality experts group on the
  validation of objective models of video quality assessment march 2000,''
  2000.

\bibitem{liu2022perceptual}
Q.~Liu, H.~Su, Z.~Duanmu, W.~Liu, and Z.~Wang, ``Perceptual quality assessment
  of colored 3d point clouds,'' \emph{IEEE Transactions on Visualization and
  Computer Graphics}, 2022.

\bibitem{sheikh2006image}
H.~R. Sheikh and A.~C. Bovik, ``Image information and visual quality,''
  \emph{IEEE Transactions on image processing}, vol.~15, no.~2, pp. 430--444,
  2006.

\bibitem{van2008visualizing}
L.~Van~der Maaten and G.~Hinton, ``Visualizing data using t-sne.''
  \emph{Journal of Machine Learning Research}, vol.~9, no.~11, 2008.

\end{thebibliography}
%



%
\begin{IEEEbiography}[{\includegraphics[width=1in,height=1.25in,clip,keepaspectratio]{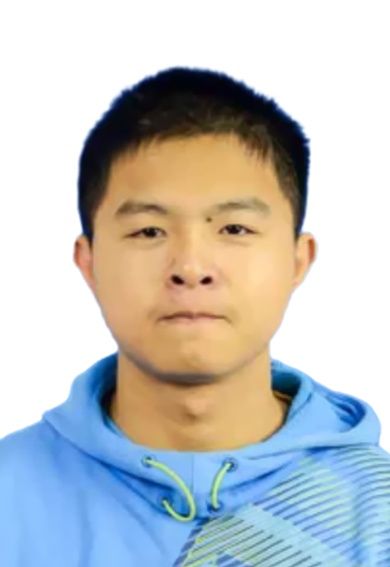}}]{Ziyu Shan} received the BS degree in electronic information and electrical engineering from Shanghai Jiao Tong University, Shanghai, China, in 2022. He is currently working toward the MS degree in information and communication engineering with Shanghai Jiao Tong University, Shanghai, China. His research interests include point cloud quality assessment and quality enhancement.
\end{IEEEbiography}

\begin{IEEEbiography}[{\includegraphics[width=1in,height=1.25in,clip,keepaspectratio]{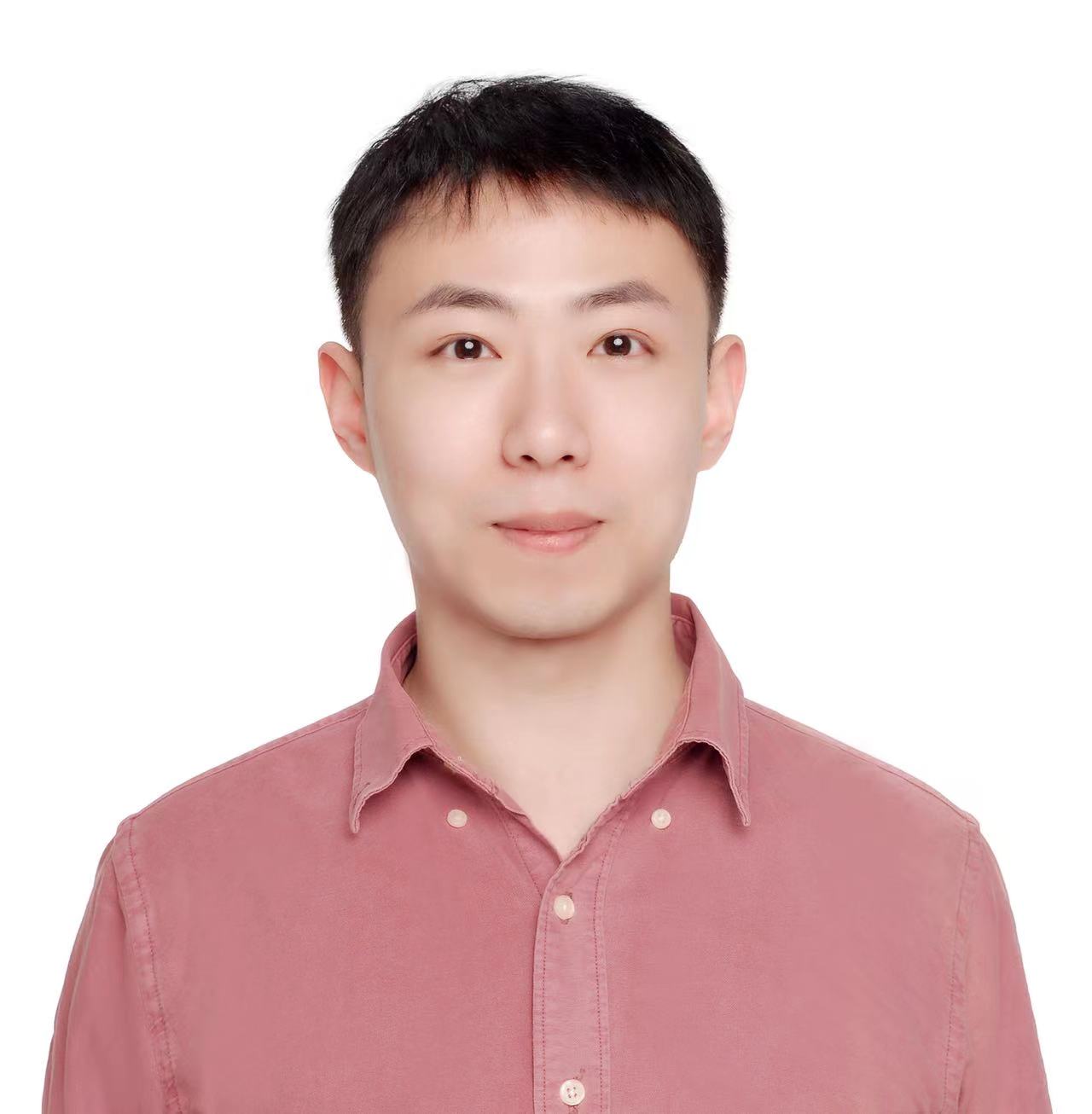}}]{Qi Yang} received the BS degree in communication engineering from Xidian University, Xi’an, China, in 2017, and the PhD degree in information and communication engineering from Shanghai Jiao Tong University, Shanghai, China, 2022. Now, he is a researcher in Tencent MediaLab. His research interests include image processing, 3D point cloud / mesh quality assessment and reconstruction.
\end{IEEEbiography}

\begin{IEEEbiography}[{\includegraphics[width=1in,height=1.25in,clip,keepaspectratio]{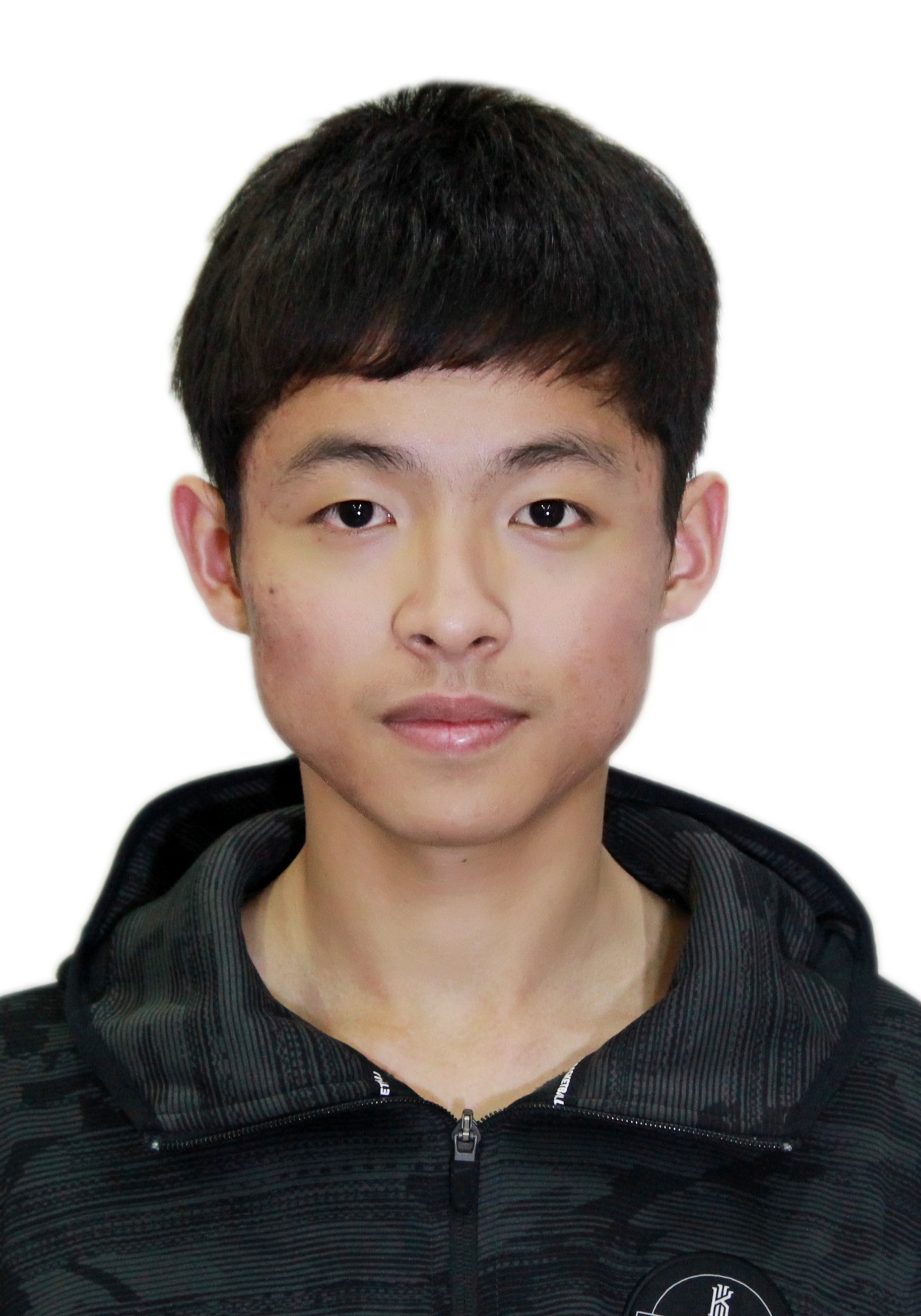}}]{Rui Ye} received the BS degree in electronic information and electrical engineering from Shanghai Jiao Tong University, Shanghai, China, in 2022. He is currently working toward the PhD degree in information and communication engineering with Shanghai Jiao Tong University, Shanghai, China. His research interests include federated learning, collaborative intelligence and image processing.
\end{IEEEbiography}

\begin{IEEEbiography}[{\includegraphics[width=1in,height=1.25in,clip,keepaspectratio]{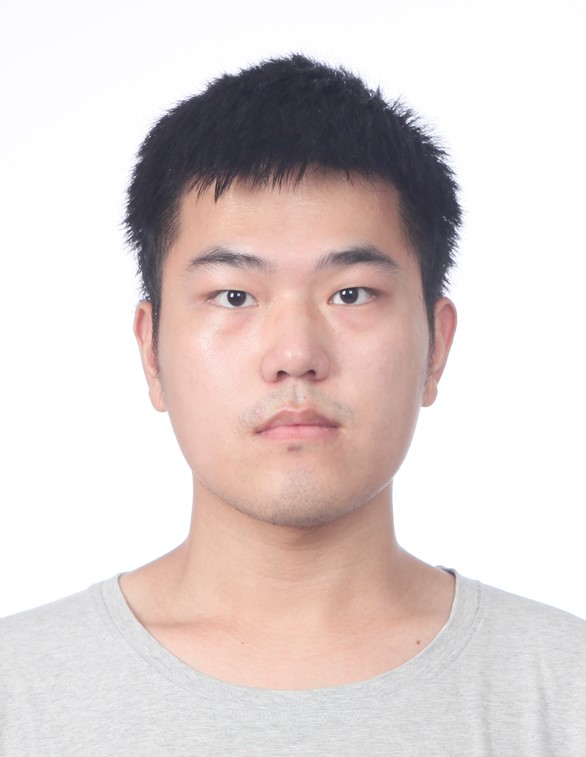}}]{Yujie Zhang} received the BS degree in electronic and information engineering from Beihang University, Beijing, China, in 2020. He is currently working toward the PhD degree in information and communication engineering with Shanghai Jiao Tong University, Shanghai, China. His research interests
include point cloud quality assessment and quality enhancement.
\end{IEEEbiography}

\begin{IEEEbiography}[{\includegraphics[width=1in,height=1.25in,clip,keepaspectratio]{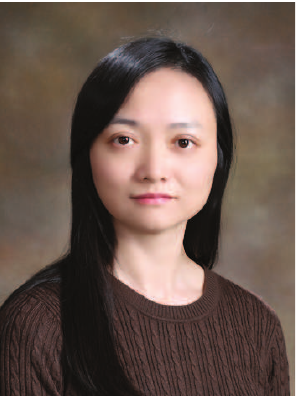}}]{Yiling Xu} (Member, IEEE) received the BS, MS,
and PhD degrees from the University of Electronic Science and Technology of China, in 1999, 2001, and 2004 respectively. From 2004 to 2013, she was a senior engineer with the Multimedia Communication Research Institute, Samsung Electronics Inc., South Korea. She joined Shanghai Jiao Tong University, where she is currently a professor in the areas of multimedia communication, 3D point cloud compression and assessment, system design, and network optimization. She is the associate editor of the IEEE Transactions on Broadcasting. She is also an active member in standard organizations, including MPEG, 3GPP, and AVS.
\end{IEEEbiography}

\begin{IEEEbiography}[{\includegraphics[width=1in,height=1.25in,clip,keepaspectratio]{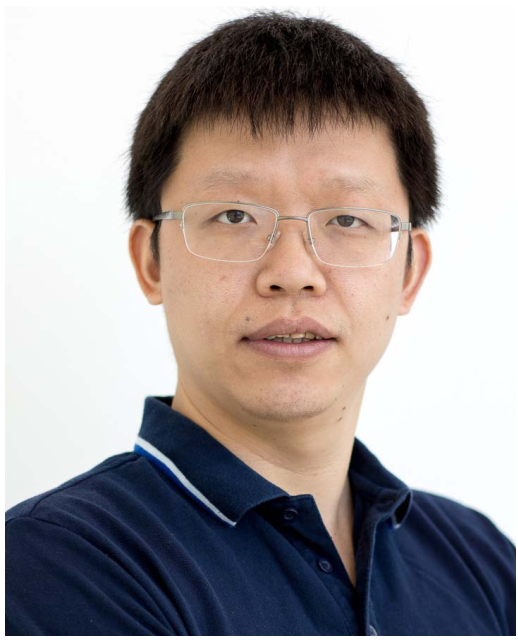}}]{Xiaozhong Xu} (Member, IEEE) received the B.S.
degree in electronics engineering from Tsinghua University, Beijing, China, the M.S. degree in electrical and computer engineering from the Polytechnic School of Engineering, New York University, New York, NY, USA, and the Ph.D. degree in electronics engineering from Tsinghua University. He has been a Principal Researcher and a Senior Manager of multimedia standards with the Tencent Media Laboratory, Palo Alto, CA, USA, since 2017. His research interests include multimedia, video and image coding, processing, and transmission. He was a recipient of the Science and Technology Award from the China Association for Standardization in 2020.
\end{IEEEbiography}

\begin{IEEEbiography}[{\includegraphics[width=1in,height=1.25in,clip,keepaspectratio]{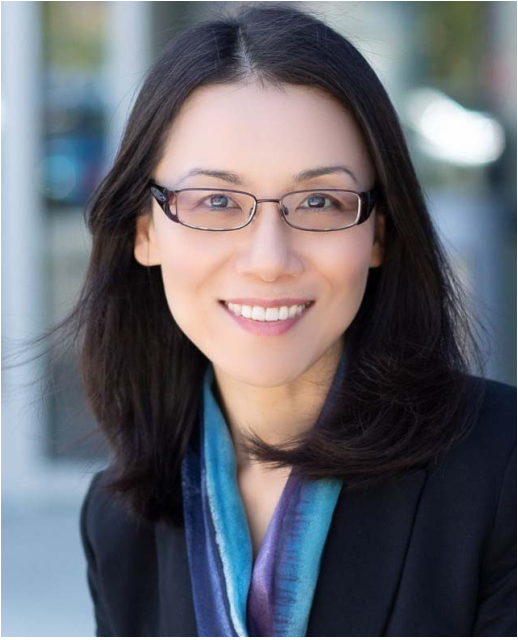}}]{Shan Liu} (Senior Member, IEEE) received the B.Eng. degree in electronic engineering from Tsinghua University and the M.S. and Ph.D. degrees in electrical engineering from the University of Southern California. She is currently a Tencent Distinguished Scientist and the General Manager of the Tencent Media Laboratory. Her research interests include audio-visual, high volume, immersive and emerging media compression, intelligence, transport, and systems. She was with the committee of Industrial Relationship of IEEE Signal Processing Society from 2014 to 2015. She served as the VP for the Industrial Relations and Development of Asia–Pacific Signal and Information Processing Association from 2016 to 2017. She was named as the APSIPA Industrial Distinguished Leader in 2018. She received the Best Associate Editor Award in 2019 and 2020. She served as a Co-Editor for H.265/HEVC SCC and H.266/VVC. She has been serving as the Vice Chair for IEEE Data Compression Standards Committee since 2019.
\end{IEEEbiography}









\end{document}